\documentclass{article}

\usepackage[preprint]{neurips_2025}

\usepackage[utf8]{inputenc} 
\usepackage[T1]{fontenc}    
\usepackage{hyperref}       
\usepackage{url}            
\usepackage{booktabs}       
\usepackage{amsfonts}       
\usepackage{nicefrac}       
\usepackage{microtype}      
\usepackage{xcolor}         
\usepackage{graphicx}

\usepackage{subfig}
\usepackage{wrapfig}
\usepackage{float}
\usepackage{amsmath} 

\usepackage{makecell}
\usepackage{xcolor}
\usepackage{multirow}
\newcommand{\xhdr}[1]{\vspace{3pt}\noindent\textbf{#1}}

\newcommand{\behaviorAI}{\text{Behavior-AI}}
\newcommand{\beliefAI}{\text{Belief-AI}}
\newcommand{\ignore}[1]{}
\newcommand{\squishlist}{
\begin{list}{{{\small{$\bullet$}}}}
{\setlength{\itemsep}{3pt}      \setlength{\parsep}{1pt}
\setlength{\topsep}{1pt}       \setlength{\partopsep}{3pt}
\setlength{\leftmargin}{1em} \setlength{\labelwidth}{1em}
\setlength{\labelsep}{0.5em} } }
\newcommand{\squishend}{  \end{list}}

\title{On the Utility of Accounting for Human Beliefs about AI Intention in Human-AI Collaboration}

\author{
Guanghui Yu$^1$, Robert Kasumba$^1$, Chien-Ju Ho$^1$,  William Yeoh$^1$ \\
$^1$Washington University in St. Louis \\
  \{guanghuiyu, rkasumba, chienju.ho, wyeoh\}@wustl.edu 
}

\begin{document}

\maketitle

\begin{abstract}
To enable effective human-AI collaboration, optimizing AI performance alone is insufficient without considering human factors. Recent research has shown that designing AI agents that account for human behavior leads to improved performance in collaborative settings. However, most existing approaches assume that human behavior remains static regardless of the AI agent’s actions. In practice, humans often adjust their behavior based on their beliefs about the AI’s intentions, that is, what they believe the AI is trying to accomplish.
In this paper, we develop collaborative AI agents that account for their human partner’s beliefs about the AI agent’s intentions and design their action plans accordingly. We first propose a model of human beliefs, extending the classical level-k reasoning framework, that captures how people interpret and reason about their AI partner’s intentions. Using this belief model, we develop an AI agent that incorporates both human behavior and human beliefs when planning its interactions.
Through extensive human-subject experiments, we show that our belief model more accurately captures human perceptions of AI intentions. Furthermore, we demonstrate that collaborative AI agents designed to account for human beliefs about their intentions significantly improve performance in human-AI collaboration.
\end{abstract}

\section{Introduction}


The potential for human-AI collaboration is immense and spans various domains, including healthcare~\citep{cheng2016risk,mobadersany2018predicting}, industrial manufacturing~\citep{bauer2008human,sherwani2020collaborative}, and workflow productivity~\citep{wilson2018collaborative}. However, despite significant improvements in AI performance over the past decade, designing AI agents capable of effectively collaborating with humans remains a challenge.

In particular, optimizing the AI system in isolation is shown to be insufficient for enhancing the performance of human-AI collaboration. To optimize collaborative performance, AI agents need to account for their human counterparts when deciding on their strategies. For example, in AI-assisted decision-making, researchers have demonstrated that, instead of optimizing AI independently, training AI to focus on improving in areas where humans typically struggle can significantly enhance the performance of human-AI teams~\citep{bansal2021most,wilder2021learning}. In human-AI collaboration, \citet{carroll2019utility} show that incorporating models of human behavior into the training of AI agents leads to higher collaborative performance compared to training the agent through self-play~\citep{silver2018general}. These studies highlight the importance of integrating human behavior into the design of collaborative AI.
However, a key limitation of these research efforts is that they mostly assume that human behavior remains static, irrespective of the actions and behavior of the AI counterpart.
In practice, humans may modify their behavior in response to their beliefs about the intentions of AI agents based on their observations of AI behavior.


\begin{wrapfigure}{o}{0.55\textwidth}
    \vspace{-10pt}
    \centering
       \null\hfill
    \subfloat[
    AI ignoring human beliefs about AI intentions.
    ]{
        \centering
        \includegraphics[width=0.46\linewidth]{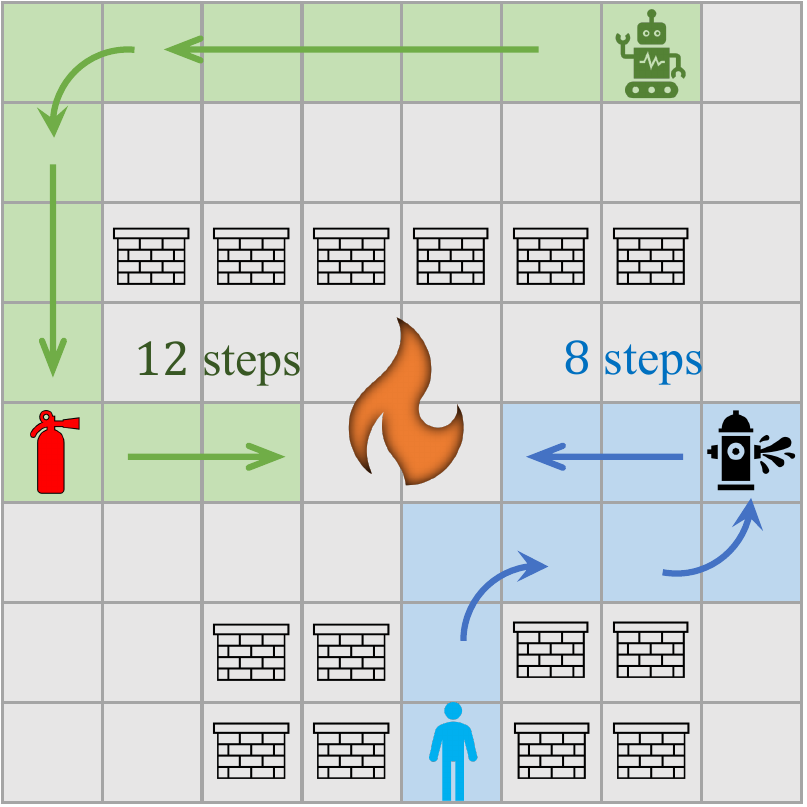}
        \label{fig:intro1}}   
        \hfill      
    \subfloat[
    AI acounting for human beliefs over AI intentions.
    ]{
        \centering
        \includegraphics[width=0.46\linewidth]{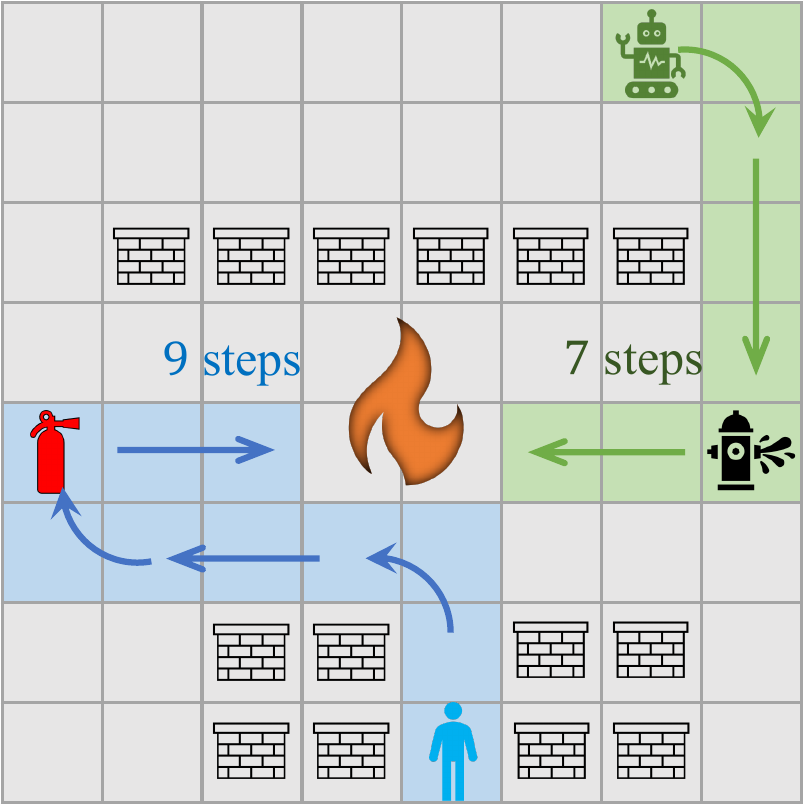}
        \label{fig:intro2}}     
           \hfill\null
    \caption{An example human-AI cooperation task. 
    }
    \vspace{-5pt}
\label{fig:example-task}
\end{wrapfigure}

In this work, we argue and aim to demonstrate that, in addition to incorporating human behavior, designing AI agents that account for human beliefs about AI intentions can further improve collaborative performance. Consider the illustrative task in Figure~\ref{fig:example-task}, where a human and an AI agent must collaborate to extinguish a fire at the center. To succeed, they must use both a fire extinguisher and a hydrant, with each retrieving one.  They cannot communicate directly, but they can observe each other’s movements.
If the AI believes the human will simply take the closest tool (the hydrant) and will not adjust their actions based on the AI’s actions, the AI would take the longer route to the fire extinguisher. However, if the human would update their behavior based on inferred AI intentions, i.e., what the AI appears to be aiming for, the AI can adopt a more efficient strategy: move toward the hydrant and let the human retrieve the fire extinguisher.
This example shows that when humans update their behavior based on their beliefs about AI intentions, an AI that accounts for those beliefs can enhance overall human-AI collaboration.
We investigate and examine the benefits of incorporating human beliefs about AI intentions in the design of AI to improve human-AI collaboration. 
Our work focuses on a setting where human and AI agents need to coordinate their goals (e.g., they need to complete different subtasks) to accomplish the overall task. 
We formulate the human-AI collaboration environment as a multi-player \emph{goal-oriented} Markov decision process (MDP).
We begin our investigation by developing models of human beliefs about AI intentions. This modeling effort extends the level-$k$ reasoning framework ~\citep{stahl1994experimental} to account for suboptimal human behavior. Specifically, we first develop a \emph{behavioral level-0} model that assumes agents take actions without considering the behavior of other agents.
We then develop the belief model by introducing a \emph{behavioral level-1} model, assuming humans interpret others' behavior as if the other agent takes actions according to the \emph{behavioral level-0} model. 

To develop and evaluate our model of human behavior (i.e., the behavioral level-0 model), we train behavioral cloning models on real-world human data and assess their performance. To build and test our model of human beliefs about AI intentions (i.e., the behavioral level-1 model), we conduct experiments where participants observe an agent’s behavior and infer its goal, allowing us to evaluate how well our model predicts human inferences. Using this belief model, we also explore designing AI policies that help humans more easily infer AI intentions from observed actions.
Overall, our results demonstrate the effectiveness of our developed models of human behavior and beliefs. 

Utilizing the developed models of human behavior and beliefs, we conduct a final set of human-subject experiments in which each human participant pairs with an AI agent in a two-player coordination game. 
We assess human-AI collaborative performance for different designs of AI agents.
We show that AI agents accounting for models of human behavior and beliefs achieve better collaborative performance when working with real-world human participants compared to AI agents that either do not consider human beliefs or disregard both human behavior and beliefs.

\xhdr{Contributions.}
The main contributions of our work are summarized as follows:
\squishlist
    
    \item We develop models of human beliefs about AI intentions, extending the level-$k$ framework to account for human suboptimality. In addition, we design AI agents capable of executing explicable action plans, making it easier for humans to infer AI intentions based on observed AI actions. 
    
    \item We design collaborative AI agents that incorporate both models of human behavior and models of human beliefs about AI intentions. To the best of our knowledge, this is among the first efforts to explicitly account for human beliefs about AI intentions in the design of collaborative AI.
    
    \item 
    Through extensive human-subject experiments with roughly 1,000 participants, we demonstrate that (1) our belief models are effective and can inform the design of AI policies that are explicable to humans, and (2) our collaborative AI design significantly outperforms state-of-the-art baselines when working with real humans. These results underscore the importance of incorporating human beliefs into collaborative AI and demonstrate the effectiveness of our approach.
\squishend

\section{Related Work}
Our work contributes to the growing body of research on human-AI collaboration~\citep{carroll2019utility,jaderberg2019human,bansal2021most,wilder2021learning}. For example, \citet{bansal2021most} show that optimizing AI in isolation can lead to suboptimal performance in human-AI collaboration. \citet{wilder2021learning} find that training AI to complement humans by excelling where humans struggle improves collaborative outcomes. \citet{carroll2019utility} demonstrate that incorporating a human model, learned from data, into AI training enhances collaboration with real users. Our work builds on this research by accounting not only for human behavior but also for human beliefs about AI intentions.

Modeling human beliefs about others' intentions has been extensively studied in the literature on level-k reasoning~\citep{stahl1994experimental,gill2016cognitive} in economics and in theory of mind~\citep{premack1978does,westby2023collective}. For example, \citet{yu2006unified} found that human behavior aligns closely with level-1 and level-2 reasoning models in cooperative games. \citet{albrecht2018autonomous} survey methods for autonomous agents modeling the beliefs and intentions of others, distinguishing between approaches for stationary versus dynamic behaviors. These models include theory of mind, recursive reasoning, plan recognition, 
and others. \citet{baker2011bayesian} propose a Bayesian theory of mind to explain how humans form joint beliefs over states in partially observable MDPs, with supporting experiments in~\citet{baker2009action, baker2017rational}. Similarly, \citet{wu2021too} apply a human modeling approach in the Overcooked experiment, decomposing the task of delivering a dish into subtasks, where players infer others' intentions and choose actions accordingly.
Our belief inference framework differs from prior theory of mind work by focusing specifically on human beliefs about AI intentions, using a more realistic human model as the basis for inferring those beliefs. A more detailed discussion of related work is provided in Appendix~\ref{sec:full-related-work}.

\section{Problem Formulation and Methods}

\subsection{Decision-Making Environment}


We formulate the human-AI cooperative decision-making environment as a multi-agent MDP, represented by $W = \langle S, \alpha, {A_{i} |_{i \in [\alpha]}}, R, P \rangle$, where $S$ is a finite set of states, $\alpha$ is a finite set of players, $A_{i}$ is the action set available to player $i$, 
$P$ is the transition function that determines the next state given all players' joint actions, 
and $R$ is the joint reward function assigned to all players given their joint actions. 
Note that in this work we consider the cooperative setting. Therefore, our formulation only incorporates a single reward function $R$ for all players, though this can be easily extended. 
We also focus on a two-player cooperative game, 
where one player is a human and the other is an AI agent. 
During decision making, neither the human nor the AI knows the other player's next action or future plans, and they cannot communicate directly. However, they can observe each other's past actions, allowing them to infer the intentions of the other player and adjust their own actions accordingly.

\xhdr{Agent intentions.}
In this work, we focus on a setting with a goal-oriented MDP, where there is a set of goals $G = \{g_1, \dots, g_k\} \subseteq S$.
The decision-making agent only receives rewards when arriving at one of the goal states, and the goal states are terminal states, i.e., $P(g|g,a) = 1~\forall g \in G, a \in A$.
Both humans and AI agents can observe past actions but not the specific goals the other agents are trying to achieve.
We use this formulation to define an agent's intention in practical terms, that is, an agent's intention refers to the specific goal the agent is aiming to achieve.
\subsection{Modeling Human Beliefs about AI Intention}
\label{sec:model}
Our model extends the level-$k$ framework~\citep{stahl1994experimental} in economics. 
In particular, we start by modeling human behavior by considering humans as level-$0$ agents who do not account for others' behavior. Unlike prior literature, we address the realistic scenario where level-$0$ agents may not behave optimally, and we call this model \emph{behavioral level-$0$} agents.
Then, to model human beliefs about others' intentions,\footnote{In this work, we sometimes use \emph{human beliefs} to refer to humans' beliefs about the intention of the AI agent.} we model humans as an extension of level-$1$ agents, who assume other agents are \emph{behavioral} level-$0$ agents\footnote{Our model can iteratively extend to higher level-$k$ agents; however, we focus on the case with $k \leq 1$.} and update their beliefs about AI intentions in a Bayesian manner based on the observation of others' behavior. More details are discribed below.

\xhdr{Modeling human behavior.}
We first describe our models of human behavior under the assumption that humans do not consider other players 
(or that they view other players as part of the environment without strategically responding).
Our approach follows standard approaches in the literature and primarily serves as a foundation for developing models of human beliefs.
In particular, a human behavior model can be represented as $H: W \rightarrow \Pi$, mapping a given environment $w \in W$ to a policy $\pi = H(w)$. We provide two examples of human behavior models used in our work below.
\squishlist
    \item \emph{Standard model.}
    First consider the standard human behavior model in MDPs, in which the goal of the human is to maximize the expected cumulative reward, and their policy only depends on the current state.
    The model can be represented by $\pi(a|s)$, indicating the probability of choosing action $a$ at state $s$.
    For the standard model that assumes decision optimality, humans choose actions maximizing the Q-function, where $Q(s,a)$ indicates the expected cumulative reward if the player takes action $a$ in state $s$ and follows policy $\pi$. $Q(s,a)$ could be calculated by standard reinforcement learning techniques such as value iteration or Q-learning.

    \item \emph{Behavioral level-$0$ model.}
    We also consider the case where we can learn human behavioral models from their historical behavior using behavioral cloning~\citep{pomerleau1988alvinn,torabi2018behavioral}. Behavioral cloning is an imitation learning approach that learns a policy from human demonstrations by mapping states to actions using supervised learning methods~\citep{bain1995framework}. We build a fully connected neural network, where the input is the state encoding and the output is the probability distribution over the action space. The model is trained using the standard gradient descent method with a cross-entropy loss.
\squishend
\xhdr{Modeling human beliefs.}
As summarized earlier, our belief models extend the ideas of level-$k$ reasoning and assume that human decision-makers perceive other agents in the environment as behavioral level-$0$ agents. 
We also refer to this belief model as \emph{behavioral level-$1$ agents}.

More specifically, we model the human belief updating process as a Bayesian inference process, which can be represented as in Equation~\eqref{eq-belief}, where $\lambda(g)$ represents the human's prior belief distributions of AI intentions, and $Pr(s_t, a_t|g)$ denotes the probability of observing $(s_t, a_t)$ given the intention (goal) $g$, according to the policy model $\pi(a|s, g)$. This policy model represents how humans perceive the actions of other players. 
If a human believes the other agent is following the standard model, then the policy is derived from the optimal policy. 
If they believe the agent follows the behavioral cloning model, the policy will reflect the learned human model.


\vspace{-10pt}
\begin{equation}
\begin{aligned}
    B(g|(s,a)_{1:t})  \propto \lambda(g) Pr((s,a)_{1:t}| g) 
                  = \lambda(g) \prod_{1:t} \pi(a_i|s_i, g) P(s_i|s_{i-1},a_{i-1}) 
\end{aligned}
\label{eq-belief}
\end{equation}

\subsection{Designing \emph{Explicable} AI Policy}
\label{sec:explicable-policy}
Prior research~\citep{ullman2009help, baker2011bayesian, zhi2020online} has demonstrated that humans often find it challenging to infer the intentions of other agents based on their behavioral traces, which also aligns with our observations in human-subject experiments (see Section~\ref{sec-exp} and the appendix for comprehensive results). Therefore, before we delve into the design of collaborative AI, we also explore whether we can leverage the model of human beliefs about AI intentions to design an \emph{explicable} AI policy, where the AI takes actions that make it easier for humans to infer its intentions.

To train an explicable AI policy, we incorporate models of human beliefs into the AI training process. Rather than solely maximizing rewards, we also align the AI policy to increase the likelihood that humans can infer its intentions from its actions, using the belief model to capture the human inference process. Specifically, we use self-play to train the models. In addition to standard MDP rewards, we introduce bonus rewards when the inferred goal, based on the belief model, matches the specified goal.
The bonus reward is set proportional to the log likelihood of the belief model inference, $\log(\pi(a|s,g))$. This encourages the AI policy to not only reach the goal but also to act in a way that makes its intentions more inferable, according to the belief model.

\subsection{Designing Collaborative AI Agents}
\label{sec:teammate-design}

We now describe how we design collaborative AI agents that account for different models of human behavior and beliefs. In the experiments, we illustrate the effectiveness of incorporating models of human beliefs by examining the performance of AI agents paired with humans through simulations and real-world human-subject experiments.

\xhdr{Training methodology.}
The main idea of our training method is through self-play. 
We incorporate the models of humans and have AI teammates play with the agents specified by the human models through simulated plays. 
We use proximal policy optimization (PPO) to train collaborative AI agents. 


\xhdr{Collaborative AI agents.}
We have designed several AI agents based on different assumptions of the human counterpart.
\squishlist
    \item \emph{Assuming humans are optimal.}
    We train an AI agent that learns to collaborate with itself through self-play. This is equivalent to assuming the human is acting optimally. 

    \item \emph{Assuming humans are behavioral level-$0$.}
    We train an AI agent that assumes the human is a behavioral level-$0$ agent, using behavioral cloning to learn the human's behavioral model. 
    \item \emph{Incorporating models of human beliefs.}
    Finally we also train an AI agent that incorporates both the models of human behavior and beliefs into the design of AI agents.
\squishend

In addition to the AI agents trained using our methodologies, we also include two state-of-the-art baselines, Fictitious Co-Play~\citep{strouse2021collaborating} and Maximum Entropy Population-based Training~\citep{zhao2023maximum}, for comparison in our experiments. 
Both algorithms develop a single best response model in response to a population of agents with diverse behaviors, thus enhancing the robustness and generality of the learned best response model.
The details of the setup and implementations are provided in the appendix.



\section{Human-Subject Experiments}
\label{sec-exp}
To evaluate our approaches, we have conducted multiple sets of human-subject experiments through recruiting close to 1,000 participants from Amazon Mechanical Turk (MTurk). The experiments are approved by the IRB of our institution. Workers were paid \$1 base payments with the potential for bonus payments. The average hourly rate is approximately $\$14$.
Given space constraints, we report only the key results. Additional results, including experiments in a varied domain and qualitative discussions of findings (e.g., the impact of environments), are provided in Appendix~\ref{sec:ori-exp}.

\subsection{Experiments Overview}
The main purpose of our experiments is to demonstrate that our proposed methodologies by incorporating human beliefs about AI intentions into the design of AI agents can significantly improve the performance of human-AI collaboration with \emph{real-world human participants}.

\begin{wrapfigure}{o}{0.45\textwidth}
    \vspace{-25pt}
    \centering
       \null\hfill
    \subfloat[Experiment 1 \& 3.]{
        \centering
        \includegraphics[width=0.45\linewidth]{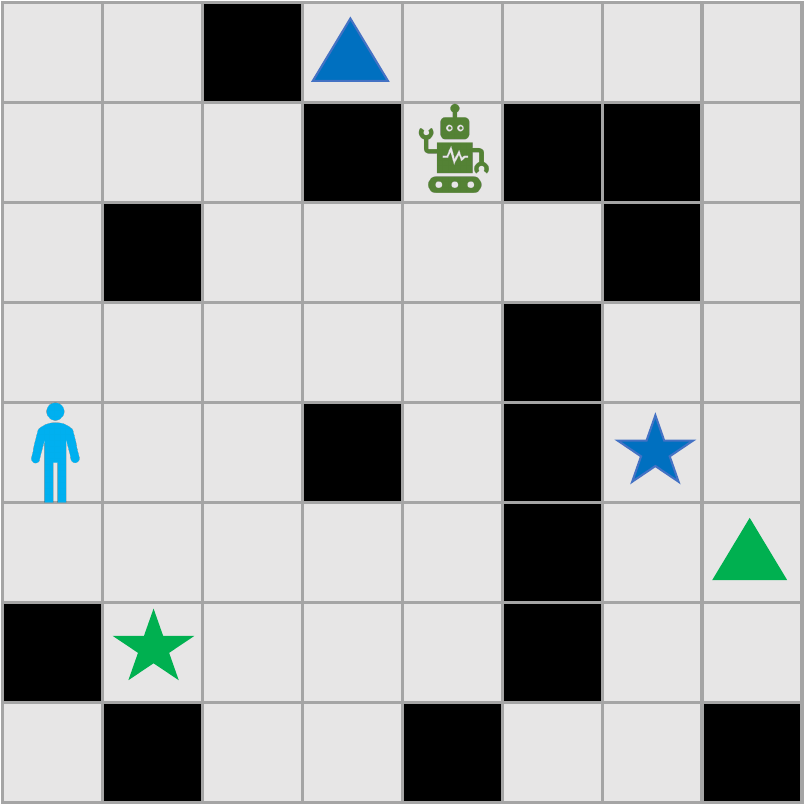}
        \label{fig:exp1}}   
        \hfill      
    \subfloat[Experiment 2.]{
        \centering
        \includegraphics[width=0.45\linewidth]{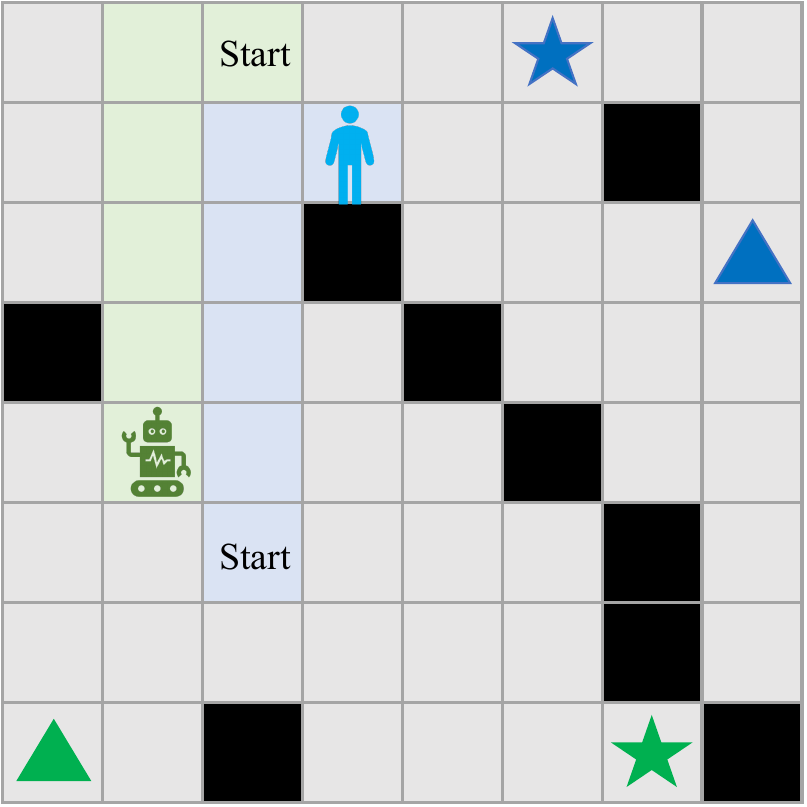}
        \label{fig:exp2}}     
           \hfill\null
    \caption{
    The interface for our human-subject experiments. In Experiment 1, participants play with an AI agent with being told the goal, focusing on reaching a goal without colliding. In Experiment 2, participants observe agent behavior traces and infer the agent's goal. In Experiment 3, participants decide their actions based on beliefs about AI behavior, without being told the goal.
    }
    \vspace{-15pt}
\label{fig:interface-all-share}
\end{wrapfigure}

\xhdr{Experiment environment.}
\label{sec-exp-setup8}
Our experiments are conducted in a grid world environment with two players and multiple goals.\footnote{We conducted experiments in two variations of the environment and
obtained qualitatively similar results in both variations. Due to space constraints, we have included the setup and results of the second variation in the main paper and moved the results of the first variation to the appendix.
}
Specifically, as shown in Figure~\ref{fig:exp1}, the grid world is 8 by 8 in size and contains the positions of both players and four possible goals. The players can choose to move \{\textit{Up, Down, Right, Left}\} or stay in their current grid. They can see each other's positions and take actions simultaneously. Each player can navigate to two of the four goals: the human player can reach one of the two goals colored blue, and the AI player can reach one of the goals colored green. When both agents reach the same type of goal (e.g., both reaching a "star" or both reaching a "triangle"), they earn positive points. However, they do not earn points if they reach different types of goals or if they collide with each other (move into the same position). We set the maximum number of actions to 20.


\xhdr{Experiment descriptions.} 
We conducted three sets of experiments, beginning with examining the efficacy of our models of human behavior and beliefs, and progressing to evaluate the performance improvements of adopting our design of collaborative AI.
In Experiment 1, we examine the effectiveness of using behavioral cloning as a model to mimic human behavior. We recruit real-world participants to play against AI agents to collect their behavioral traces and then evaluate the prediction accuracy of using behavioral cloning to model human behavior.
In Experiment 2, we examine whether we can leverage the model of human beliefs to design an explicable AI policy (as described in Section~\ref{sec:explicable-policy}). Specifically, we recruit real-world participants, provide them with behavioral traces generated by different AI policies, and evaluate whether the AI policies that account for human beliefs produce behavioral traces that make it easier for humans to infer AI intentions.
Finally, in Experiment 3, we examine whether the collaborative AI that accounts for human beliefs lead to higher performance in human-AI collaboration. Again, we recruit real-world human participants, pair them with different AI agents (as described in Section~\ref{sec:teammate-design}), and measure the overall team performance.


\subsection{Experiment 1: Evaluating Behavior Models}

In our first experiment, our goal is to examine the effectiveness of utilizing behavioral cloning to model human behavior. 
Note that this approach has been proposed and examined in various domains~\citep{carroll2019utility}. 
This experiment is a replication of prior works in our setting. 
The main purpose is to examine whether this method works well in our setting, as our proposed approaches in developing human belief models and collaborative AI are built on top of models of human behavior. 

\xhdr{Experiment setup.}
We recruited 190 workers from Amazon Mechanical Turk. Each participant was asked to play 30 navigation games within a grid world (as shown in Figure~\ref{fig:exp1}) alongside an AI model. The goal of this experiment was to develop a model of human behavior that predicts a sequence of actions given a specified goal. To this end, we instructed participants on which goal to reach in each round and recorded their behavior as the collected dataset.

\xhdr{Experiment results.}
We divided the collected data of human actions into three sets: training, validation, and testing. The training set comprised data from 152 workers, including approximately 136,000 instances of user decisions, while the validation and testing sets each contained data from 19 workers, amounting to around 17,000 instances of user decisions each. We employed a 4-layer Multilayer Perceptron (MLP) model, where the input is the current environment layout, and the output predicts the next human action. We fine-tuned the hyperparameters, such as learning rate, hidden layer size, and L2 penalty, based on validation errors.

\begin{table*}[h]
    \caption{The prediction accuracy for human behavior was evaluated assuming optimal behavior and using a data-driven model in Experiment 1. 
    }
    \centering
    \begin{tabular}{lccc}
    \hline
     & Training Accuracy & Validation Accuracy & Testing Accuracy \\ \hline
    Assuming Optimal Behavior & 0.4498 & 0.4327 & 0.4459 \\
    Data-Driven Model & 0.8547 & 0.7831 & 0.7899 \\ \hline
    \end{tabular}
    \label{tab:share-env-acc}
\end{table*}

We compared the performance of our learned model with a model predicated on optimal agent behavior, defined as taking the shortest path to the goal. The training, validation, and test accuracies of both models are presented in Table~\ref{tab:share-env-acc}. These results clearly reveal that human behavior deviates significantly from the assumed optimality. This deviation highlights the importance of incorporating a realistic model of human behavior into human-AI cooperation frameworks. Moreover, the data-driven model trained using behavioral cloning achieves reasonable and much higher predictive performance.

\subsection{Experiment 2: Evaluating Belief Models}
We next evaluate our model of human beliefs. 
In particular, we examine whether the belief model can enable us to develop explicable AI policy (as described in Section~\ref{sec:explicable-policy}), making it easier for humans to infer AI intention based on AI actions.\footnote{We also evaluated the accuracy of our belief model by comparing its predictions with users’ directly reported inferences about others’ behavior. As shown in Appendix~\ref{sec:appendix-belief}, our model outperforms baseline predictors of human beliefs. However, the results also reveal that human beliefs are highly noisy. When shown a randomly selected behavior trace, human goal inference is often close to random guessing. Consequently, we shift the experiment’s focus to whether AI agents can plan their actions to make their goals easier for humans to infer.}

\xhdr{Experiment setup.}
In this experiment, each recruited participant is presented with the behavioral traces from both agents in the environment, as shown in Figure~\ref{fig:exp2}. Each behavioral trace is a sequence of actions generated by an AI policy for a player. Participants are then asked to infer the intention of each agent, i.e., the goal each agent is trying to reach. If participants can better infer the goal from the behavior generated by a certain AI policy, it indicates that the AI policy is more \emph{explicable} in that it generates actions that make it easier for humans to infer AI intentions.


\xhdr{AI policy.}
We now describe the set of AI policies that are used for evaluations. 
As baselines, the first two AI policies are not intended to be explicable. 

\squishlist
    \item \emph{Inexplicable-Optimal-L0}: This AI takes the shortest path towards the goal.
    \item \emph{Inexplicable-Behavioral-L0}: This AI follows the behavioral model learned in Experiment 1.
\squishend

We then include two AI policies that take actions intended to be explicable. To design AI policies to be explicable, we need to account for human belief models in the design of AI policies, as the AI will try to take actions that make it easier for humans to infer the goal, assuming models of human beliefs that describe how humans make inferences. The two belief models we account for are: (1) the classical level-$1$ model, where the human assumes the other agent is taking optimal level-$0$ actions, and (2) the behavioral level-$1$ model, where the human assumes the other agent is taking behavioral level-$0$ actions as described in Section~\ref{sec:model}. Coupled with these two belief models, and using the methodology described in Section~\ref{sec:explicable-policy}, we train the following two explicable AI policies.

\squishlist
    \item \emph{Explicable-Optimal-L1}: This AI policy is trained to be explicable but assumes that humans belief model is the classical level-$1$ model, where they assume the other agent is optimal.
    \item \emph{Explicable-Behavioral-L1}: This AI policy is trained to be explicable but assumes that human belief model is the behavioral level-$1$ model, where they assume the other agent is behavioral level-$0$.
\squishend

\begin{wrapfigure}{R}{0.5\textwidth}
    \vspace{-15pt}
    \centering
        \includegraphics[width=0.98\linewidth]{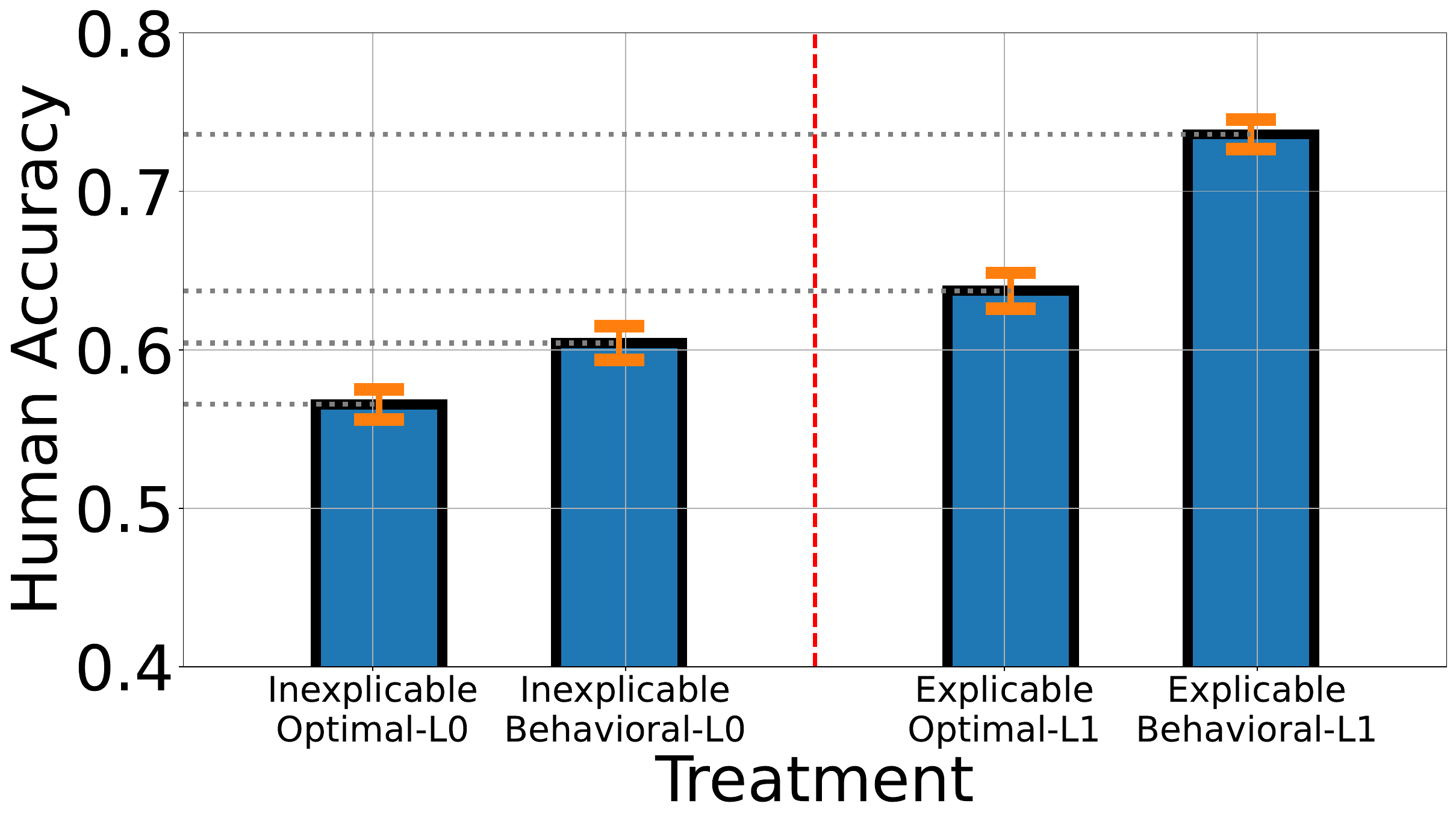}
    \caption{The results of Experiment 2. Using our approach to design explicable AI leads to higher human accuracy in inferring AI intentions. Combined with our belief model, it results in an AI policy that makes it easiest to infer AI intentions.} 
    \vspace{-15pt}
\label{fig:belief-infer-share-human}
\end{wrapfigure}

\xhdr{Experiment results.}
We recruited 300 workers from Amazon Mechanical Turk, randomly assigning them to one of four treatments, where each treatment represents showing participants traces generated by one of the AI policies above.
Each participant was tasked with identifying the goals of the player in 30 different scenarios. The length of actions is drawn from the range 4 to 8. 
To incentivize effort, participants were awarded a $\$0.03$ bonus for each correctly identified goal.

The experiment results are shown in Figure~\ref{fig:belief-infer-share-human}.
First, we observe that when the AI is not attempting to be explicable, humans generally struggle to infer AI intentions, with the accuracy of correctly inferring the goal for both inexplicable policies being around 55\% to 60\%, only slightly better than random guessing. However, if we use our methodology from Section~\ref{sec:explicable-policy} to train the AI policy to be explicable, it significantly improves the accuracy of humans in inferring AI intentions. Moreover, when the training is coupled with our belief models, the accuracy for humans to infer AI intentions is much higher than all other policies, demonstrating the effectiveness of both our belief models and the methodologies in training explicable AI policy.

\subsection{Experiment 3: Evaluating Collaborative AI}
Finally, as the main experiment, we evaluate our collaborative AI design. The objective is to show that incorporating human beliefs into the design of collaborative AI can enhance human-AI collaborative performance 
through both simulations and human-subject experiments.


\xhdr{Collaborative AI design.}
As described in Section~\ref{sec:teammate-design}, we train collaborative AI agents using three different corresponding human models. 
\squishlist
    \item \emph{Self-Play}: the collaborative AI agent that is trained assuming they are playing with itself. 
    \item \emph{\behaviorAI}: the collaborative AI agent that is trained assuming the human partner is the behavioral level-$0$ agent. This corresponds to the existing approaches of accounting for human behavior in AI design~\cite{carroll2019utility}.
    \item \emph{\beliefAI}: the collaborative AI agent that is trained assuming the human partner is the behavioral level-$1$ agent, i.e., it accounts for human belief models into the design of AI.
\squishend
In addition to the three collaborative AI agents trained using our methodology, we also include two state-of-the-art baselines for comparison. These methods leverage zero-shot population-based training (PBT) techniques. While they do not explicitly account for specific models of human behavior, they are designed to be robust in collaborating with different teammate behaviors.
\squishlist
    \item \emph{FCP}: Fictitious Co-Play is a variant of fictitious play proposed by \citet{strouse2021collaborating}. It consists of two stages: in the first stage, a population of agents is trained independently using different random seeds and checkpoints; in the second stage, a model is trained as the best response to the agent population.
In our experiment, we select the model trained with 8 partners.
    \item \emph{MEP}: Maximum Entropy Population-based training is proposed by ~\citet{zhao2023maximum}. The agent population is trained using population entropy bonus to increase the diversity of agents, then a single best response model is trained with respect to the agent population. In our experiment, we select the model trained with 8 partners.
\squishend
We emphasize again that this set of comparisons spans a wide range of state-of-the-art baselines in the literature on collaborative AI design, including standard multi-agent RL approaches (e.g., self-play), methods that account for human behavior~\citep{carroll2019utility} (\behaviorAI), and zero-shot population-based approaches (\emph{FCP} and \emph{MEP}). Our approach is distinct in that it incorporates a model of human belief into the design of collaborative AI---to our knowledge, the first to do so.


\xhdr{Simulations.}
We first run simulations to evaluate the performance of different pairings of human models and collaborative AI designs, where we assume the human agent is optimal (\emph{self-play}), an behavioral level-$0$ agent (\emph{Behavioral Model}), and behavioral level-$1$ agent (\emph{Behavior\&Belief}).  The evaluation is based on $10,000$ randomly generated environments which filter out cases where the distance between the two goals for the same player is smaller than 3 to allow models to adjust their behavior based on the inference of their teammate's actions.

We partner the collaborative AI agents with the three simulated human agents and measure the collaborative performance. The simulation results are shown in Table~\ref{tab:share-env-team}.
For both the FCP and MEP baselines, despite the absence of human data in their training processes, they outperform the self-play AI when paired with the human behavior model, demonstrating their adaptability to working with teammates exhibiting unknown behavior. However, we also observe that incorporating the appropriate human behavior model directly into the training process can further enhance human-AI collaborative performance, as demonstrated by the superior performance of Behavior-AI when paired with the behavior model. Finally, when working with a behavioral level-1 agent that attempts to infer AI intentions, our proposed Belief-AI stands out, significantly outperforming other AI agents.

\begin{table*}[h]
\caption{Simulation results of collaborative performance over 10,000 testing cases in Experiment 3. The column players represent different AI agents, and the row players represent different simulated human models. The results demonstrate that while the zero-shot methods (FCP and MEP) show stable performance, when human agents infer AI intentions to act accordingly, designing AI that accounts for human beliefs brings significant performance improvements.}
\centering
    \begin{tabular}{@{}lccccc@{}}
    \hline
    \multirow{2}{*}{Human Model} & \multicolumn{5}{c}{AI Agent} \\ \cline{2-6} 
       & Self-Play & FCP & MEP & \behaviorAI & \beliefAI  \\ \hline
    Self-play & $\textbf{0.6245}$ & 0.6136 & 0.6131 & 0.5250 & 0.6177 \\
    Behavior Model & 0.4902 & 0.5782 & 0.5927 & \textbf{0.6334} & 0.5755 \\
    Behavior\&Belief & 0.6411 & 0.6255 & 0.6484 & 0.6574 & \textbf{0.7675} \\ 
    \hline
    \end{tabular}
    \label{tab:share-env-team}
\end{table*}

These results highlight the importance of incorporating appropriate human models (of behavior and beliefs) for training collaborative AI. Furthermore, under conditions where humans attempt to infer AI intentions and act accordingly, designing AI that accounts for human beliefs can significantly improve human-AI collaborative performance.


\begin{wrapfigure}{R}{0.5\textwidth}
    \centering
    \vspace{-15pt}
    \includegraphics[width=0.98\linewidth]{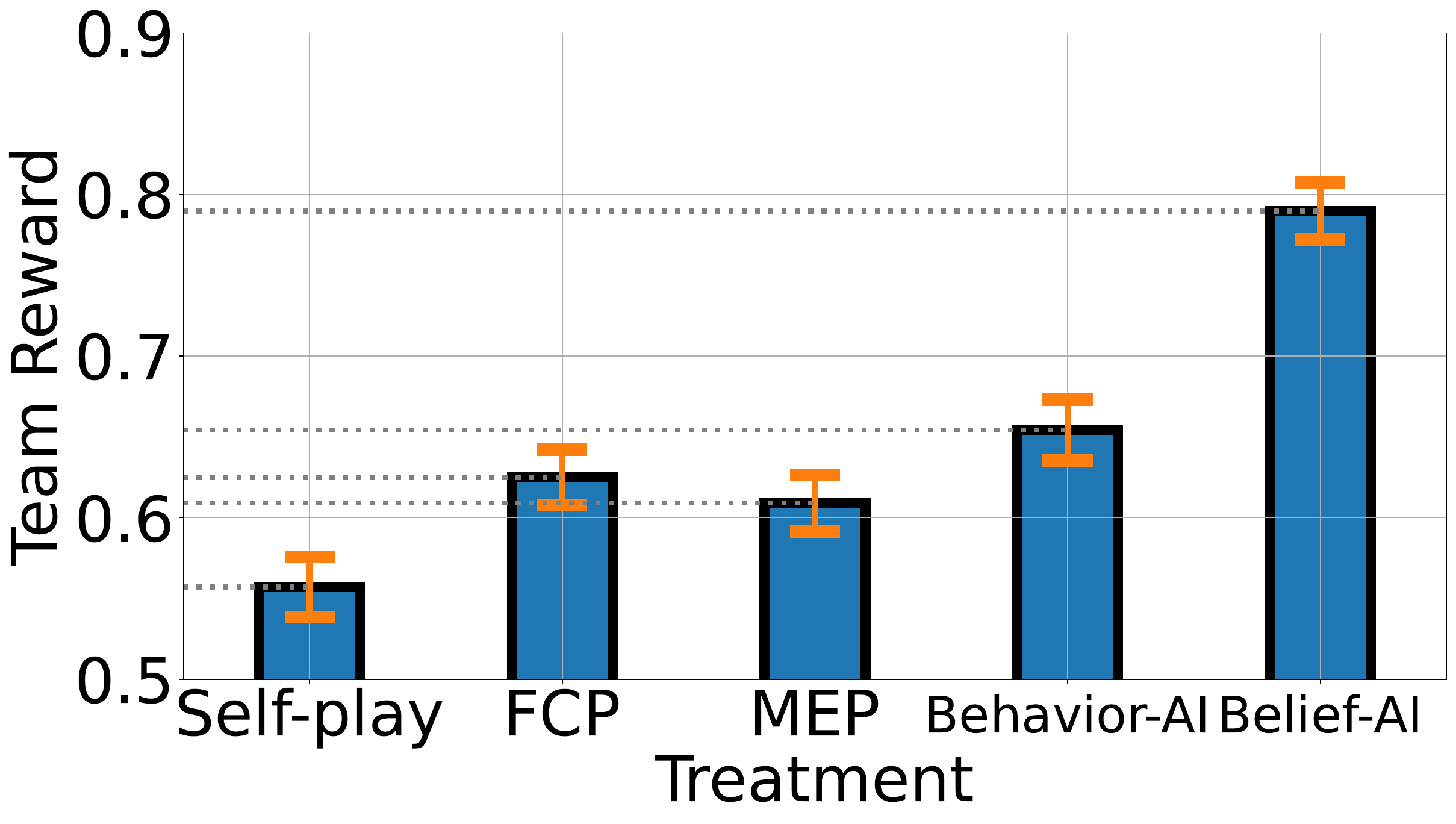}
    \caption{The human-AI collaborative performance in Experiment 3. Designing collaborative AI that accounts for human beliefs (Belief-AI treatment) significantly outperforms all other baselines when paired with real humans.}

    \vspace{-10pt}
\label{fig:human-ai-shareenv}
\end{wrapfigure}

\xhdr{Human-subject experiments.}  
While the simulation results are promising, the significant performance improvement from incorporating human beliefs into AI design is based on the assumption that humans follow our belief model. To examine whether this performance improvement carries over to collaborations with real humans, we recruited 400 workers and randomly assigned them into five treatment groups, each interacting with a different AI design. Participants were tasked with playing 30 games, preceded by three tutorial games designed to familiarize them with the gameplay. Participants could receive a \$0.05 bonus payment for reaching the same type of goal as the AI agent in each of the 30 tasks.

We first examine whether the data-driven model learned in Experiment 1 still makes good prediction on human behavior in Experiment 3 (i.e., whether the learned behavior model performance generalizeds to new experiment). 
Overall, we find that the model achieves a accuracy of $73.19\%$ in predicting human behavior in Experiment 3 across all treatments, demonstrating the generalizability of learned human behavior model.
More detailed discussion is provided in Appendix~\ref{appendix-bc-exp}.

We then examine the collaborative performance for pairing human participants with different AI design.
The experiment results are shown in Figure~\ref{fig:human-ai-shareenv}. 
They demonstrate that the collaborative AI trained with models of human beliefs achieved the highest collaborative performance when working with real humans, significantly outperforming other AI designs. This outcome not only validates our AI design but also suggests that real human actions align with our belief models in this environment.

More formally, statistical analysis shows significant differences in the performance of AI models paired with humans, with $p<0.0001$ for comparisons between self-play AI and \behaviorAI, and $p<0.0001$ for comparisons between \behaviorAI versus \beliefAI.
The differences between MEP and FCP, as well as between FCP and \behaviorAI, are not significant ($p>0.1$). The difference between self-play AI and any other treatments is significant ($p<0.0001$). 

\section{Conclusion and Discussion}
\label{sec-conclusion}
Our work investigates whether accounting for human beliefs about AI intentions can improve human-AI collaboration.
We make several key contributions. First, we extend level-k frameworks to develop human belief models that are empirically more accurate. Second, we introduce general methodologies for training collaborative AI agents that incorporate these models—the first to do so, to our knowledge. Finally, through extensive human-subject experiments, we show that AI agents designed with an understanding of human beliefs about AI intentions lead to improved collaborative performance with real-world users.
This underscores the importance of considering human beliefs in the design of collaborative AI and demonstrates the effectiveness of both our belief model and agent design.
Overall, our work points to a promising direction for building AI systems that more effectively align with their human partners.

\xhdr{Generalization and limitations.} 
Our approach has proven effective through human-subject experiments in grid world environments of varying complexity. However, like most human-subject studies, the findings are limited to specific settings. To improve external validity, future work should evaluate the approach in more realistic domains, such as applications involving physical robots or complex multi-agent systems. 
Additionally, our approach relies on historical data to model human behavior and beliefs, assuming these remain stable over time.  This may not hold true in real-world scenarios where human inference capability can change in response to prolonged exposure to AI or due to evolving task contexts. Future research could focus on developing adaptive models that can learn and update human behavior and belief patterns in real-time to address this limitation. Investigating the scalability of these models to larger, more diverse datasets and environments is also important and interesting for assessing the generalizability and robustness of our approach.

\section*{Acknowledgments}
This work is supported in part by a J.P. Morgan Faculty Research Award and a Global Incubator Seed Grant from McDonnell International Scholars Academy.

\bibliographystyle{named}
\bibliography{cite}


\clearpage
\appendix

\section{Related Work}
Our work contributes to the growing body of research on human-AI collaboration~\citep{carroll2019utility,jaderberg2019human,bansal2021most,wilder2021learning, myers2024learning}. For example, \citet{bansal2021most} demonstrate that optimizing AI in isolation may lead to suboptimal performance in human-AI collaboration, compared to training AI with consideration for human performance. \citet{wilder2021learning} show that training AI systems to complement humans by performing better in areas where humans struggle results in improved collaborative performance. \citet{carroll2019utility} illustrate that incorporating a human model, learned from human data, into the training of AI enhances performance when these AI systems collaborate with real humans.
Our work extends this line of research by not only accounting for human behavior when designing AI but also incorporating human beliefs about AI intentions.

While the approaches in the above works often assume knowledge or data of humans, there have also been zero-shot coordination techniques being proposed to train cooperative AI agents when human data are not available.  
For example, \citet{jaderberg2019human} use population-based reinforcement learning to improve the robustness of trained AI agents. 
\citet{strouse2021collaborating} develop a variety of self-play agents by varying random seeds of neural network initialization and various checkpoints, and train a best response model tailored to the agent population, discovering that humans prefer to partner with this model.
\citet{zhao2023maximum} utilize the advances of maximum entropy reinforcement learning to encourage the diversity of agent populations and show the learned AI models could adapt to real humans. 
This line of work aims to design collaborative AI that is robust to diverse human behavior. However, they do not explicitly account for human beliefs over AI intentions during human-AI collaboration. 
In this work, we have included these approaches as baselines and show that accounting for human beliefs leads to better performance in our setting.

At a high level, our work aligns with the broader effort to understand and model humans in computational systems~\citep{choudhury2019utility, shah2019feasibility, kwon2020humans, chan2021human, reddy2021assisted, narayanan2022does, narayanan2023does, treiman2023humans, treiman2024consequences, treiman2025people}. 
More technically, for example, inverse reinforcement learning\citep{ng2000algorithms, abbeel2004apprenticeship, ramachandran2007bayesian} aims to infer reward functions in Markov decision processes (MDPs) by observing demonstrations of the optimal policy. When the demonstrator is human, these demonstrations may be noisy or exhibit behavioral biases. 
Several studies\citep{evans2016learning, shah2019feasibility, hughes2020inferring, zhi2020online} have addressed this by incorporating human behavioral biases into the inference process, aiming to infer both the rewards and biases simultaneously.
Imitation Learning~\citep{hussein2017imitation} also seeks to develop models that can mimic human behavior based on demonstrations.
Recently, there has been a growing amount of research that incorporates human models into computational and machine learning frameworks~\citep{KO-14, tang2019bandit, tang2021bayesian, egr2021, egr2021-2, tang2021linear, yu2022environment, yu2023encoding, kasumba2024data, feng2024rationality, hong2024learning}.

From the perspective of modeling human beliefs about others' intentions, this has been discussed in the literature on level-$k$ reasoning~\citep{stahl1994experimental,gill2016cognitive} in economics and in the theory of mind~\citep{premack1978does,chen2021visual, westby2023collective, li2023theory}. For instance, \citet{yu2006unified} found that human behavior aligns closely with level-$1$ and level-$2$ reasoning in cooperative games. \citet{albrecht2018autonomous} survey works on agents modeling the beliefs and intentions of other agents, distinguishing between methods for modeling stationary or changing agent behaviors.
\citet{bara2021mindcraft} models human common ground in cooperative Minecraft tasks by leveraging theory of mind and develops AI models to infer human beliefs. \citet{baker2011bayesian} propose a Bayesian theory of mind to describe how humans model joint beliefs of state in a partially observable MDP, conducting human experiments in the works of~\citet{baker2009action, baker2017rational}. Similarly, \citet{wu2021too} apply a human modeling approach in the Overcooked experiment, breaking down the task of delivering a dish into several subtasks, such as picking up an ingredient or placing an ingredient into a pot, where the player infers others' intentions regarding these subtasks and selects their own accordingly before taking actions. 
Our work extends this line of research in two ways.  First, our belief inference framework differs from previous theory of mind literature by modeling and incorporating human beliefs about AI intentions specifically, using a more realistic human model as the basis for inferring these intentions. Second, and more importantly, beyond modeling human beliefs over others' intentions, our work focuses on the design of collaborative AI that account for human beliefs.

\label{sec:full-related-work}

\section{Additional Experiments}
\label{sec:ori-exp}

Besides experiments in Section~\ref{sec-exp}, we also conduct additional experiments in a simple grid world environment with multiple goals to illustrate our study of humans' beliefs about AI behavior. 
We also explore how to improve the model accuracy of behavioral cloning approach.

\subsection{Experiment Environment: Grid World with Multiple Goals}

We use grid worlds of size of $6\times 6$ in both simulations and human experiments. 
Similar to the environment setup in Section~\ref{sec-exp-setup8}, the grid world contains a start position, two goal positions, and some blocked positions that the player cannot enter. The player needs to move from the start position towards one of the goal positions. The player can choose to move \{Up, Down, Right, Left\}.
The player will get a positive reward upon reaching the goal, and we set the maximum number of actions to be 20. 

We first recruit participants to engage in a single-player game and record their behavior. We then build human behavior model using behavioral cloning. To evaluate our proposed approaches in modeling human beliefs and designing cooperative AI, we conducted additional two sets of experiments. In the second set of experiments, we provide participants with different traces of actions from other agents, again in a single-player game, and ask participants to infer the goal of the agents. This experiment helps us evaluate whether our belief model leads to better predictions of human beliefs over others' behavior. Afterwards, we conduct a two-player game in the third experiment, where humans are paired with different AI agents to examine the team performance of our design of collaborative AI.

\begin{figure}[h]
    \centering
    \subfloat[Experiment 4.]{
        \includegraphics[width=0.4\linewidth]{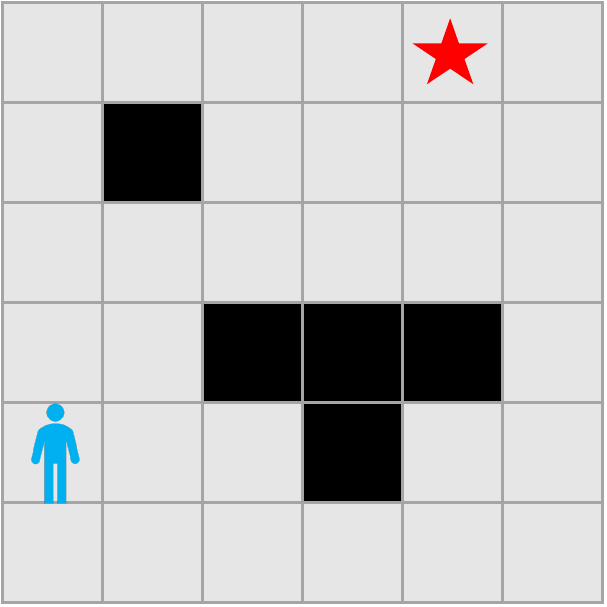}
        \label{fig:exp4}}         \hfill
    \subfloat[Experiment 5.]{
        \includegraphics[width=0.4\linewidth]{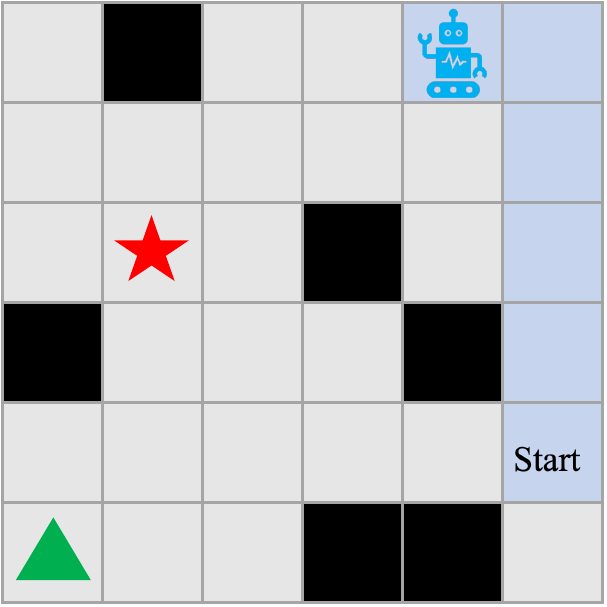}
        \label{fig:exp5}}      
        \\
    \subfloat[Experiment 6.]{
        \centering
        \includegraphics[width=0.98\linewidth]{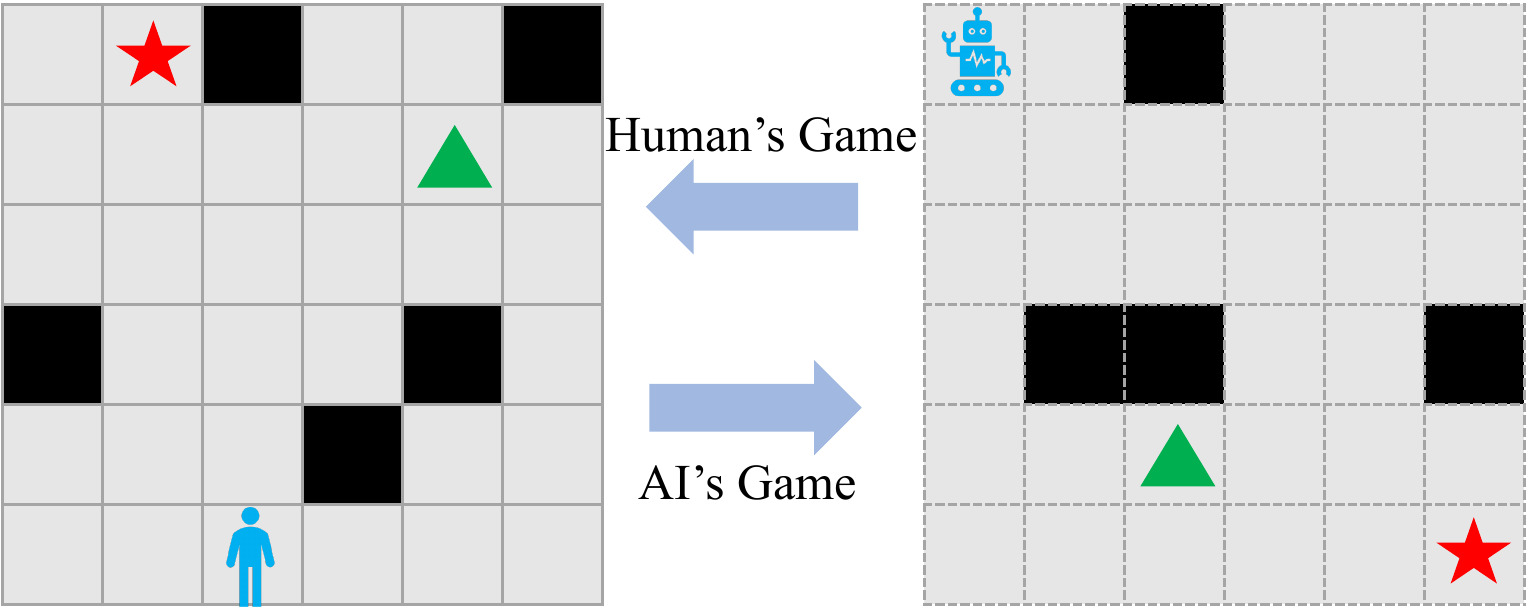}
        \label{fig:exp6}}
    \caption{Human-subject experiment interfaces. In Experiment 4, each participant is asked to control the player to move to the goal (red star). In Experiment 5, each participant is provided a trace of the behavior by another agent, and is asked to infer which goal the agent is trying to reach. In Experiment~6, each participant is playing with an AI agent in separate environments. The participants only receive bonus rewards by reaching the same goal (red star or green triangle) as the AI agent.}
    \label{fig:interface-all6}
\end{figure}

The interfaces of our experiments can be seen in Figure~\ref{fig:exp4}, \ref{fig:exp5}, and~\ref{fig:exp6}. The detailed descriptions of the experiment and interfaces are included in Appendix~\ref{sec-human-exp}.

\subsection{Experiment 4: Evaluating Behavior Models in Single-Player MDP}

Similar to Experiment 1, we recruited 200 workers from Amazon Mechanical Turk.
Each recruited worker was asked to play 15 navigation games within a grid world, as shown in Figure~\ref{fig:exp4}. Our goal is to leverage the collected data to create a data-driven model of human behavior. We divided the collected data of human actions into three sets: training, validation, and testing. The training set comprised data from 160 workers, including approximately 70,000 instances of user decisions, while the validation and testing sets each contained data from 20 workers, amounting to around 8,800 instances of user decisions each. 
The model and parameter tuning were similar to what we did in Experiment 1.

\xhdr{Evaluation of data-driven behavior models.}
The training, validation, and test accuracies of assuming optimal behavior and behavioral models are presented in Table~\ref{tab:data-driven-acc}. Results show that data-driven model could predict human behavior more accurately than assuming humans are optimal.

\begin{table*}[!ht]
\caption{The prediction accuracy for human behavior for different human models in Experiment 4.}
\centering
\begin{tabular}{lccc}
\hline
 & Training Accuracy & Validation Accuracy & Testing Accuracy \\ \hline
Assuming Optimal Behavior & 0.7266 & 0.6964 & 0.7131 \\
Data-Driven Model & 0.9189 & 0.8136 & 0.8422 \\ \hline
\end{tabular}
\label{tab:data-driven-acc}
\end{table*}

\subsection{Experiment 5: Evaluating Human Beliefs}
\label{sec:appendix-belief}

We developed human belief models (behavioral level-1) similar to previous experiments. We examine whether belief inference using behavioral model aligns with real humans, and whether we can design AI behavior such that it is easier for humans to infer the goal of the AI agent.

\xhdr{Experiment setup.}
The experiment setup is presented in Figure~\ref{fig:exp5}. 
The grid world contains a starting position and two goal positions. For each participant, we show them a trace of behavior from another agent and ask the participant to infer which of the two goal the agent is trying to reach.


\xhdr{Experiment 5.1: Examining the belief models.}
We recruited 200 workers from Amazon Mechanical Turk to compare standard level-$1$ and behavioral level-$1$ model. Each worker was asked to infer the goal for 25 behavioral traces. We compared the performance of two belief models and the worker accuracy, as shown in Table~\ref{tab-belief}. As we can see from the table, humans are generally poor at inferring the goal of other agents: their accuracy only reaches $59.37\%$ in inferring the goal of the other agent. Moreover, the behavioral level-$1$ model, which accounts for the human behavior model in the belief model, captures human beliefs better than the standard level-$1$ model, which assumes human behavior is optimal. 

\begin{wrapfigure}[11]{R}{0.5\textwidth}
    \centering
    \vspace{-10pt}
    \includegraphics[width=0.95\linewidth]{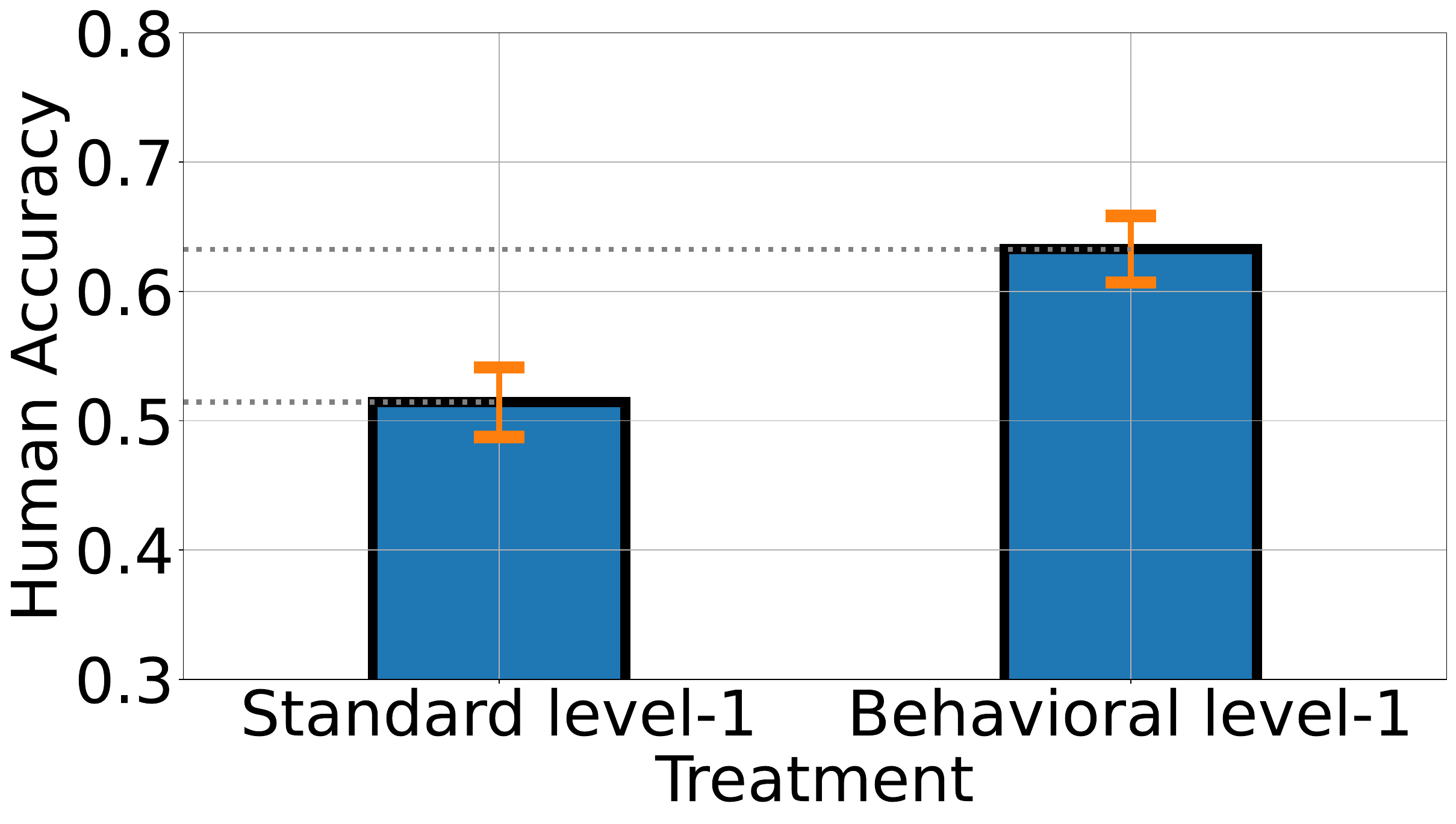}
    \caption{Human evaluations of belief inference accuracy regrading AI goals.}
\label{fig:belief-human}
\end{wrapfigure}

\begin{table*}[h]
    \caption{Performance comparison between Bayesian inference framework assuming standard model and human behavior model. }
    \centering
    \begin{tabular}{lcccc}
        \hline
     & Consistency to Human Predictions & Cross Entropy Loss  \\ \hline
    Standard level-$1$ & 0.4977 & 0.9284  \\
    Behavioral level-$1$ & 0.5764 & 0.7506  \\ 
    True goals & 0.5937 & 0.6631 \\ \hline
    \end{tabular}
    \label{tab-belief}
\end{table*}

\xhdr{Experiment 5.2: Developing explicable AI policy.}
As demonstrated in Experiment 5.1, humans generally struggle to infer the goals of other agents based on others' behavior. 
To design AI agents which makes inference of their goals easier by humans,
we train AI agents to maximize the likelihood that humans can accurately infer these goals from their behavior. 



We recruited 400 workers from Amazon Mechanical Turk, randomly assigning them to one of two treatments, to assess the behavior of AI agents.
Each participant was tasked with identifying the goals of the player in 30 different scenarios. 
Participants were awarded a $\$0.03$ bonus for each correctly identified goal.
Figure~\ref{fig:belief-human} displays the results of human evaluation. The findings indicate that humans achieve higher accuracy in inferring the correct goals of AI agents when employing Behavioral level-$1$ model.

\subsection{Experiment 6: Evaluating Collaborative AI }
Utilizing developed models of human behavior and beliefs, we follow the same methodology in Experiment 3:  train different collaborative AI, pair them with
different human models, and examine the collaborative performance in simulations and human subject experiments (via recruiting 300 workers).

The setup of our Experiment 6 is shown in Figure~\ref{fig:exp6}. 
The goal for the human-AI team is for both agents to reach the same goal in their own environments (both reaching ``red star'' or both reaching ``green triangle'') within a time limit. The team will not get points if they reach different goals or one of the players fails to reach any goal. 

\begin{wrapfigure}{R}{0.5\textwidth}
    \centering
    \includegraphics[width=0.95\linewidth]{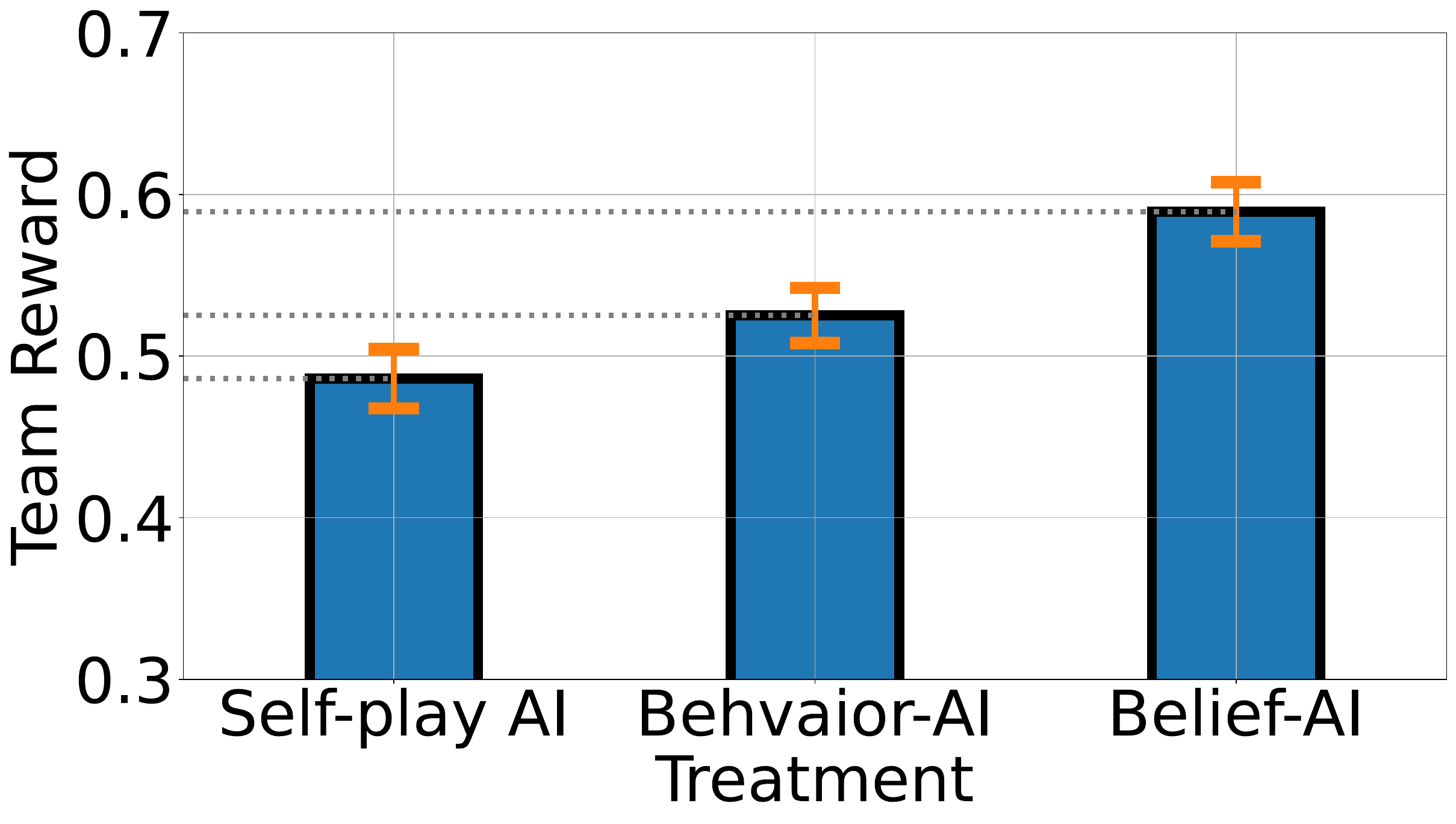}
        \caption{Average collaborative reward in  experiment 6. }
        \label{fig:human-ai}
\end{wrapfigure}

Simulations results are shown in Table~\ref{tab:team-sim}. Simulations indicate that collaborative performance is highest when an AI agent is paired with the human model used to train it. 
Figure~\ref{fig:human-ai} presents the average collaborative reward in human-subject experiments. Statistical analysis revealed significant differences in the performance of AI models when paired with human participants, with $p$-values of $0.0166$ for comparisons between the treatments \emph{self-play} and \emph{Behavior-AI}, and $p<0.0001$ for \emph{Behavior-AI} versus \emph{Belief-AI}. These results demonstrate that incorporating human beliefs into the deign of AI agents enhances collaborative performance when working with real humans.

\begin{table}
    \caption{Simulation results of human-AI collaborative rewards in Experiment 6. Columns players are different AI agents, and row players are different simulated human models.}
    \centering
    \begin{tabular}{@{}lccc@{}}
    \hline
    \multirow{2}{*}{Human Model} & \multicolumn{3}{c}{AI Agent} \\ \cline{2-4} 
       & Self-play & \behaviorAI & \beliefAI  \\ \hline
    Self-play AI & \textbf{0.7828} & 0.4780 & 0.6164 \\
    Behavior  & 0.5926 & \textbf{0.7584} & 0.4552 \\
    Behavior\&Belief & 0.6919 & 0.7268 & \textbf{0.7813} \\ \hline
    \end{tabular}
    \label{tab:team-sim}
\end{table}

\subsection{Extensive Results of Behavioral Cloning}
\label{appendix-bc-exp}

\begin{figure}[h]
    \centering
    \subfloat[Behavioral cloning models.]{
        \centering
        \includegraphics[width=0.48\linewidth]{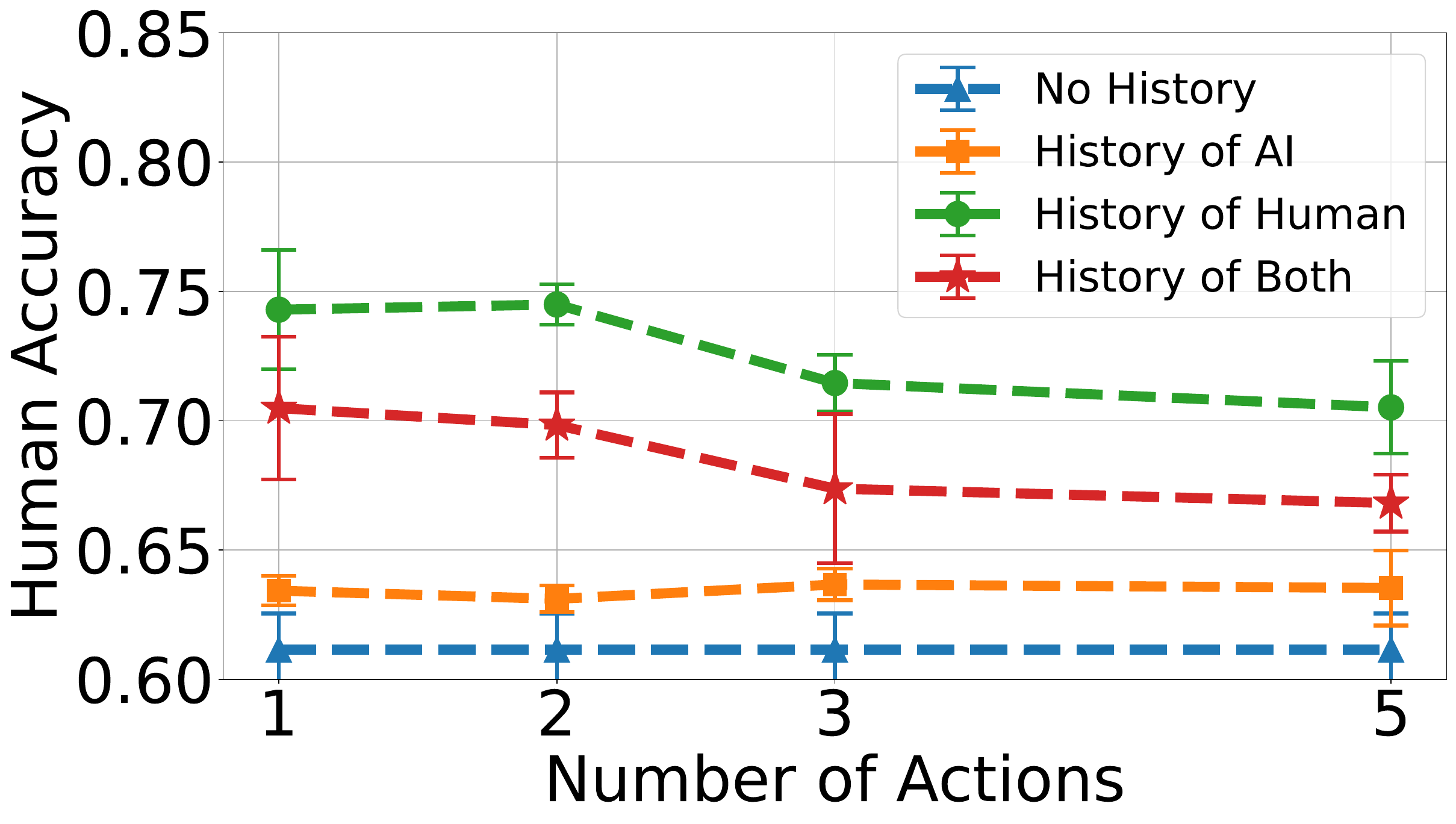}
        \label{fig:bc}}
        \hfill
    \subfloat[Behavioral cloning \& belief models.]{
        \centering
        \includegraphics[width=0.48\linewidth]{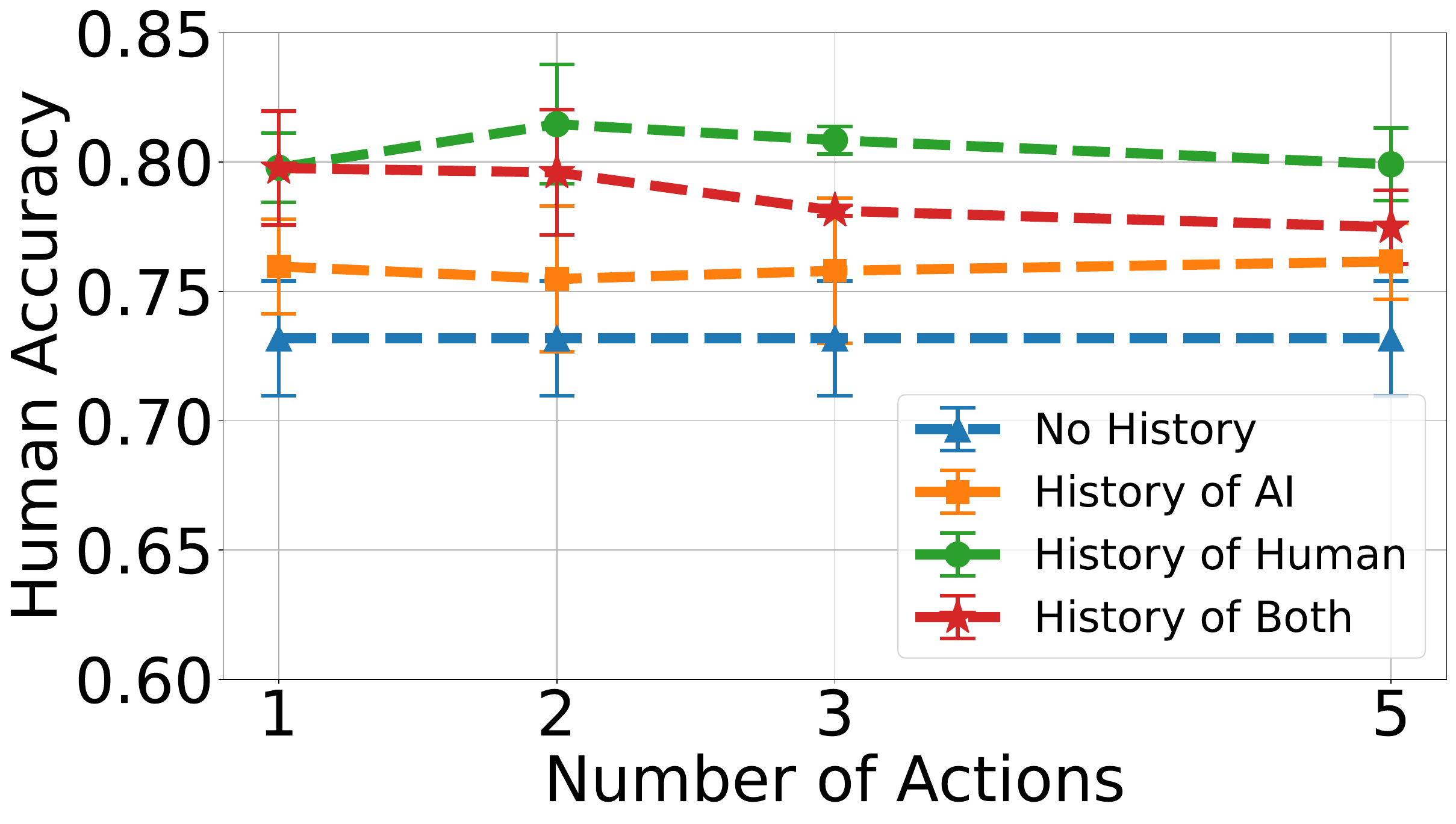}
        \label{fig:bc-belief}}      
    \caption{Evaluation of model accuracy on human-subject experiment data through incorporation of action history.}
    \label{fig:BC-history}
\end{figure}

We investigate techniques to enhance the accuracy of behavioral cloning models. Our experiment results reveal that human actions depend significantly on historical actions. We explicitly include the action history of both AI and human players as part of the model input to determine if this improves the prediction of human actions. The findings, presented in Figure~\ref{fig:BC-history}, indicate that incorporating action history enhances the accuracy of our behavioral cloning model predictions. Additionally, our belief model consistently boosts prediction accuracy across nearly all scenarios.

In Experiment 1, we aggregate all data to train a behavioral cloning model. We also train models on human data collected with different AI teammates and evaluate them across treatments, observing consistent average accuracy ranging from $75.4\%$ to $78.9\%$. Since no significant differences were detected, we aggregate all the data to create a single human model.  We also observed suboptimal performance from human participants in Experiment 1. Even when human participants are provided with suggested goals during data collection, there is only a $55.0\%$ chance that both players will reach the same type of goals across all treatments. Besides, there is a notable chance of two players colliding (about $15.8\%$) or ending in different types of goals (about $25.6\%$). 

\begin{figure}[h]
    \centering
       \null\hfill
    \subfloat[Layout where Belief-AI (11 participants) and Self-play AI (9 participants) perform similarly.]{
        \centering
        \includegraphics[width=0.45\linewidth]{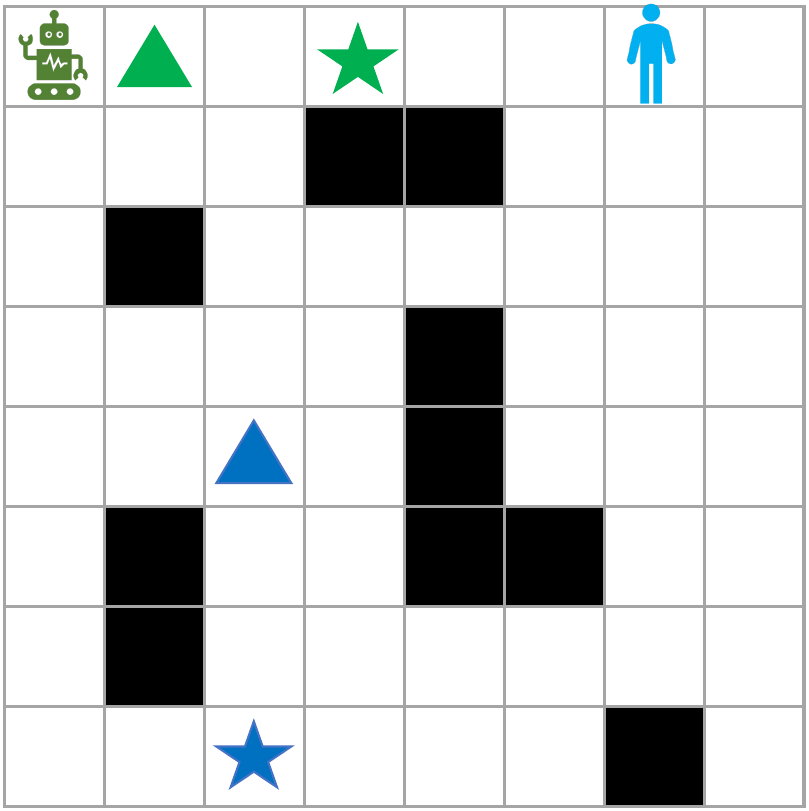}
        \label{fig:bad-exmaple}}   
        \hfill      
    \subfloat[Layout where Belief-AI (12 participants) significantly outperforms Self-play AI (13 participants).]{
        \centering
        \includegraphics[width=0.45\linewidth]{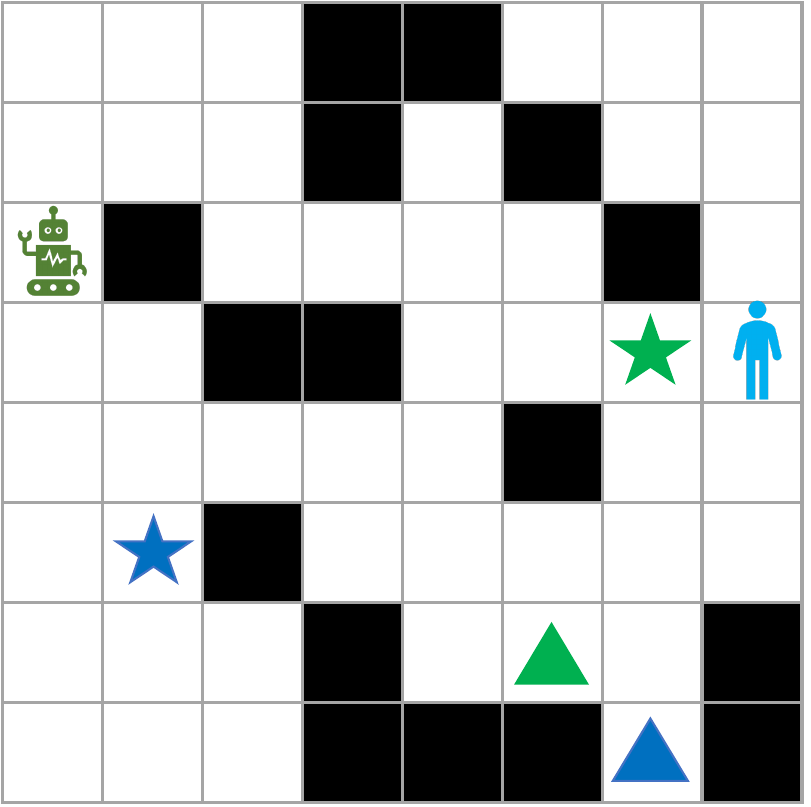}
        \label{fig:good-example}}     
           \hfill\null
    \caption{
Examples of game layouts highlighting variations in performance across treatments.
    }
\label{fig:layout_examples}
\end{figure}

We also evaluate the model’s consistency accuracy in Experiment 3 and across different environment layouts. The data-driven model with a random goal achieves an accuracy of $61.15\%$, while incorporating belief inference improves accuracy to $73.19\%$ across all treatments, consistent with the trend shown in Figure~\ref{fig:BC-history}.
While the environment layout influences model accuracy, on average, the behavioral cloning model aligns with human behavior significantly better than assuming an optimal model. Figure~\ref{fig:layout_examples} highlights scenarios where performance varies across treatments in Experiment 3. In the layout shown in Figure~\ref{fig:bad-exmaple}, most AI types remain in their starting position and act only after the human player shows a clear preference for their goal. Here, the performance of Self-play AI is almost identical to Belief-AI, with no significant differences across other treatments. However, in the layout shown in Figure~\ref{fig:good-example}, Belief-AI moves upward to signal its preference for the star, resulting in 10 out of 12 human players also reaching the star. In contrast, Self-play AI moves downward without clearly indicating its goal, leading to only 5 out of 13 participants aligning with the AI's goal.

\section{Implementation Details}
\label{sec-implement}


All our experiments are conducted on a cluster of 40 CPU cores (Intel Xeon Gold 6148 CPU $@$ 2.40GHz), 2 GPUs (NVIDIA Tesla V100 SXM2 32GB), and a maximal memory of 80GB. For completeness, we provide the details of the implementation and the tuning of hyperparameters below. 

\xhdr{State encoding of grid world.} 
For a grid world of size $N \times N$ and contains $k$ different objects, we use $k$ channels to encode the environment state. Each channel is a binary matrix of size $N \times N$, indicating whether it contains specific objects. For example, in a single-player environment with one goal (environment in Experiment 4), we encode space, blocks, the start position, and goal positions as four channels. Thus the input state is of size $4 \times 6 \times 6$ before flattening. The state encoding is used as input of our behavioral cloning models and AI agents.

\xhdr{Behavioral cloning for grid world of size $6\times6$.} 
We construct a neural network with four fully connected layers to model human behavior. The input is the state encoding, represented as a matrix of dimensions $4 \times 6 \times 6$, and the output is the probability of taking various actions. The network parameters are initialized using the Glorot uniform initializer. We employ the Adam optimizer and cross-entropy loss to optimize the neural network parameters. The batch size is set to $1024$, and the maximum number of training epochs is set to $1000$. Training is halted early if the validation loss begins to increase. The hyperparameters are adjusted based on the performance on the validation dataset. The number of nodes for each hidden layer is tuned within the range $\{64, 128, 256, 512, 1024\}$, the initial learning rate within the range $\{0.001, 0.003, 0.01, 0.03, 0.1\}$, and the L2 penalty within the range $\{0, 0.01, 0.03, 0.1\}$. Each training process takes $10$ to $30$ minutes depends on hyperparameters. In our reported results, we choose hidden layer size of $512$, learning rate $0.01$ and L2 penalty $0.01$. 

\xhdr{Behavioral cloning for grid world of size $8\times8$.} 
The neural network structure mirrors that used for the grid world of size $6 \times 6$. The input is the state encoding, represented as a matrix with dimensions $8 \times 8 \times 8$, and the output is the probability of taking various actions. The initialization, optimizer, loss function, batch size and training process are consistent with those used in the grid world of size $6 \times 6$. The maximum of training epochs is set to $3000$.
The hyperparameters are adjusted based on the performance on the validation dataset. The number of nodes for each hidden layer is tuned within the range $\{1024, 2048, 4096\}$, the initial learning rate within the range $\{0.001, 0.003, 0.01, 0.03, 0.1, 0.3\}$, and the L2 penalty within the range $\{0, 0.01, 0.1\}$. In our reported results, we choose hidden layer size of $2048$, learning rate $0.01$ and L2 penalty $0.01$.


\xhdr{AI agent training for grid world of size $6\times6$.}
We utilize the Proximal Policy Optimization (PPO) and actor-critic framework for our AI model. Both the actor and critic networks employ four fully connected layers, maintained at the same size. The input is the state encoding of a two-player game, which includes two game states and the index vector (the input vector length is $290$). The critic network outputs a single value, while the actor network produces a distribution over the action space. The batch size is set to $256$, and the maximum number of training episodes is limited to $1000$. The discount factor is set to $0.99$. For PPO optimization, the clipping parameter $\epsilon$ is fixed at $0.2$, and the number of PPO epochs is $10$. The number of nodes for each hidden layer is tuned within the range $\{64, 128, 256, 512, 1024\}$, and the initial learning rate is adjusted within the range of $\{0.001, 0.003, 0.01, 0.03, 0.1\}$. Each training process takes about $2$ to $6$ hours depends on hyperparameters. In comparisons between the self-play AI and the AI trained to play with a human model, we use the same model size (hidden layer size of $256$) and the same initial learning rate of $0.01$.

\begin{wrapfigure}[13]{R}{0.5\textwidth}
    \centering
    \includegraphics[width=1\linewidth]{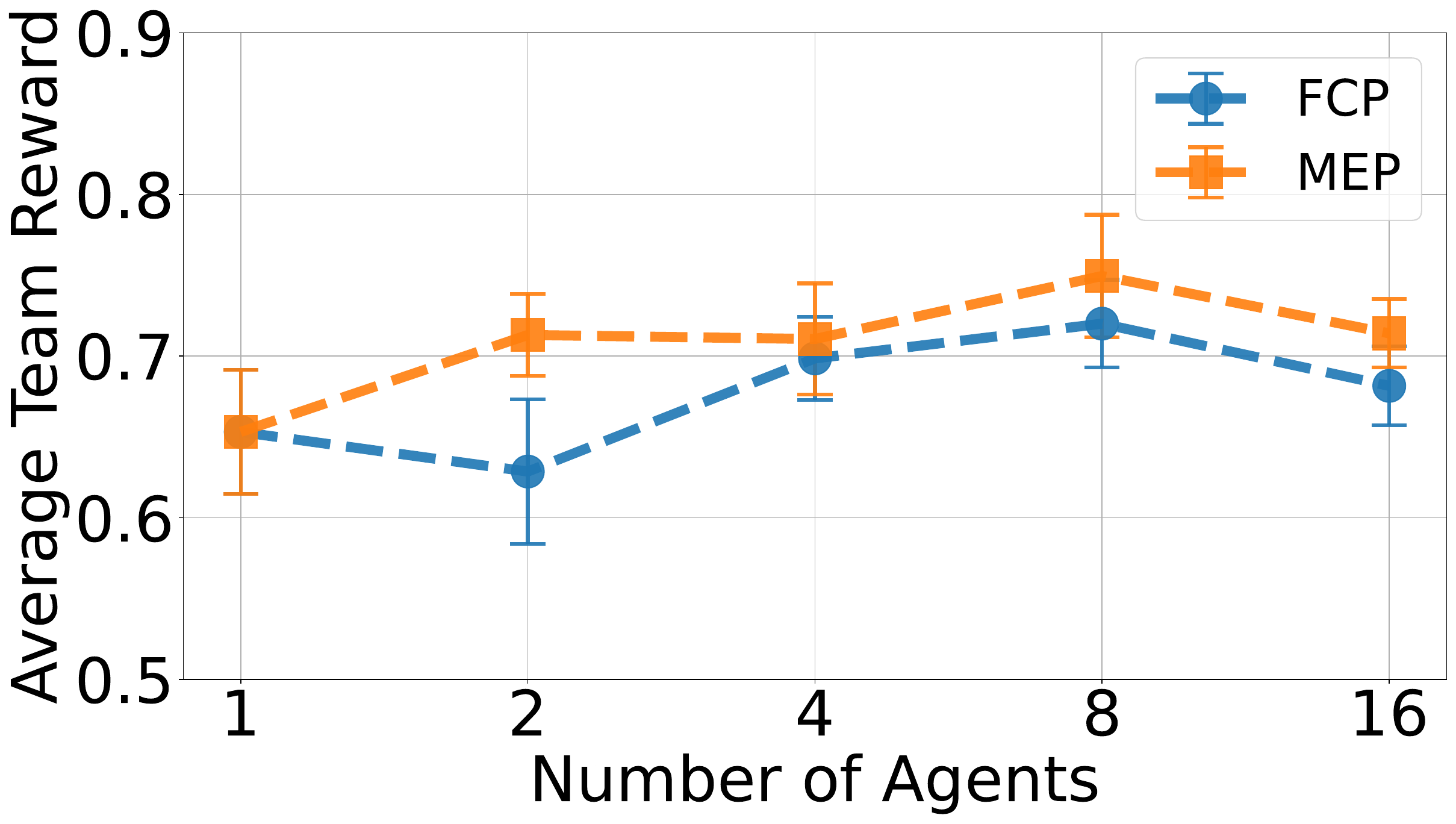}
        \caption{Average team reward for models trained across different population sizes using the FCP and MEP algorithms. }
        \label{fig:pbt-agents}
\end{wrapfigure}

\xhdr{AI agent training for grid world of size $8\times8$.}
The model structure mirrors that of the grid world of size $6 \times 6$. The input size is adjusted to accommodate the new state representation (the input vector length is $514$).
The training batch size, discounting factor, maximum number of training epochs and clipping parameters remain unchanged.
The number of nodes for each hidden layer is tuned within the range $\{256, 512, 1024, 2048\}$, and the initial learning rate is adjusted within the range of $\{0.01, 0.03, 0.1, 0.3, 0.5\}$. In comparisons between the self-play AI and the AI trained to play with a human model, we consistently use the same model size (hidden layer size of $512$) and the same initial learning rate of $0.1$.

\xhdr{Selection of population size.}
For FCP and MEP algorithms used in Section~\ref{sec-exp}, the population size will directly influence the model performance. We vary the number of agents in our simulations as shown in Figure~\ref{fig:pbt-agents}. The model trained with 8 partners is chosen for use in further simulations and human-subject experiments for comparative analysis.

\section{Human Experiment Details}
\label{sec-human-exp}

\begin{figure*}[t]
    \centering
    \subfloat[Experiment 1 \& 3 interface: humans play games with AI agents in grid worlds of size $8\times 8$.]{
        \includegraphics[width=0.48\linewidth]{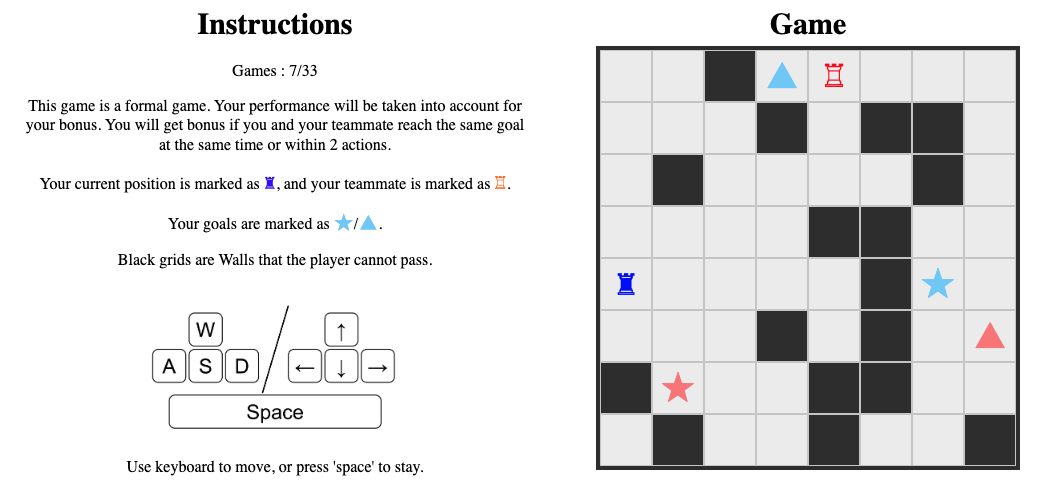}
        }   \hfill  
    \subfloat[Experiment 2 interface: humans infer goals of other players in grid worlds of size $8\times 8$.]{
        \includegraphics[width=0.48\linewidth]{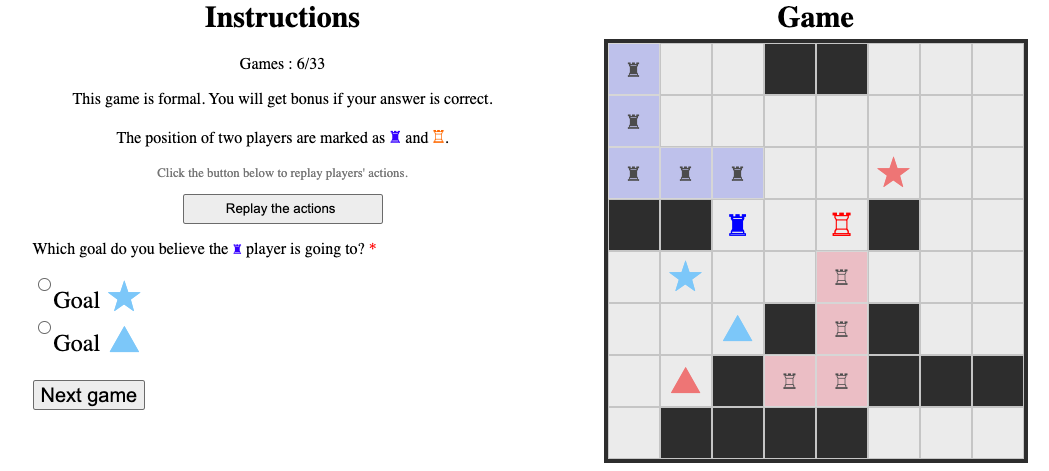}
        }   \hfill
    \subfloat[Experiment 4 interface: humans play single-player navigation games in grid worlds of size $6\times 6$.]{
        \includegraphics[width=0.48\linewidth]{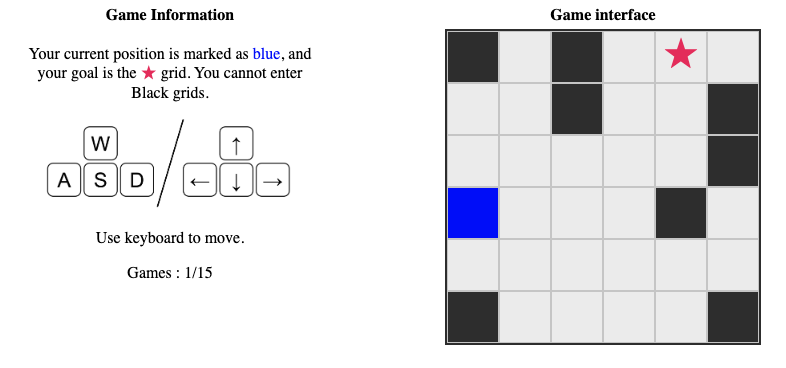}
        }
        \hfill
    \subfloat[Experiment 5 interface: humans infer goals of other players in grid worlds of size $6\times 6$.]{
        \includegraphics[width=0.48\linewidth]{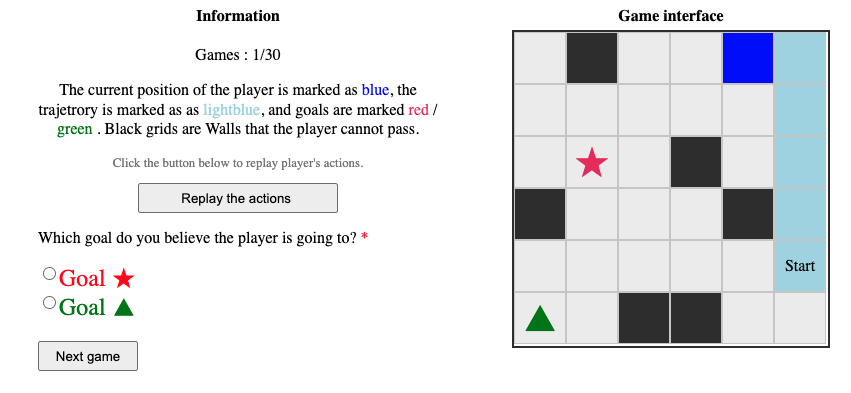}
        }   \hfill     
    \subfloat[Experiment 6 interface: humans play games with AI agents in grid worlds of size $6\times 6$.]{
        \includegraphics[width=1.0\linewidth]{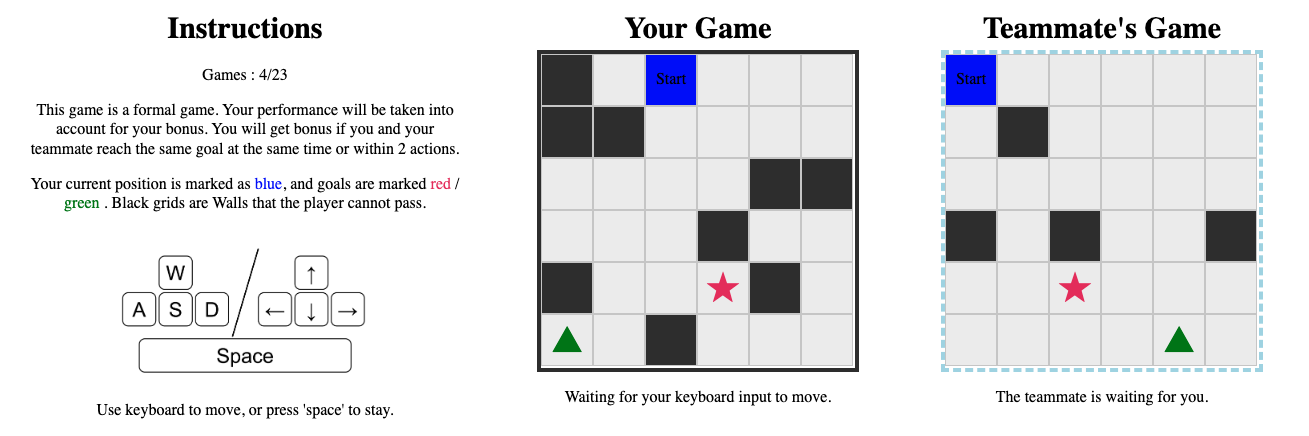}
        } \hfill
    \caption{Human-subject experiment interfaces. }
    \label{fig:interface-all}
\end{figure*}

We include more details about human-subject experiments. The experiment interface is shown in Figure \ref{fig:interface-all}. We conduct four human-subject experiments, and recruited 1990 workers from Amazon Mechanical Turk in total. Table~\ref{tab-worker-count} lists the number of workers, the number of tasks per worker and the number of treatments in each experiment. Table~\ref{tab-demo} contains the demographic information of all the workers.

\begin{table*}[h]
    \caption{Number of workers in each human-subject experiment.}
        \label{tab-worker-count}
    \centering
    \begin{tabular}{lccccc}
    \hline
 & Workers & Tasks per Worker & Treatments & Base Payment & Bonus per Task \\ \hline
Experiment 1 & 190 & 30 & 3 & \$1.00 & \$0.05 \\
Experiment 2 & 300 & 30 & 4 & \$1.00 & \$0.03 \\ 
Experiment 3 & 400 & 30 & 5 & \$1.00 & \$0.05 \\ 
Experiment 4 & 200 & 15 & 1 & \$1.00 & 0 \\
Experiment 5.1 & 200 & 25 & 1 & \$1.00 & 0 \\
Experiment 5.2 & 400 & 30 & 2 & \$1.00 & \$0.03 \\
Experiment 6 & 300 & 20 & 3 & \$1.50 & \$0.05 \\ 
\hline
    \end{tabular}
\end{table*}

\begin{table*}[h]
\centering
\caption{Demographic information of all the participants in our experiments.}
\begin{tabular}{lcc}
    \hline
     Group & Category & Number \\
    \hline
    \multirow{4}{*}{Age }
    &20 to 29  &714\\
     &30 to 39  &974\\
     &40 to 49  &176\\
     &50 or older  &72\\
     &Other & 54\\
    \hline
    \multirow{3}{*}{Gender }
    &Female&639\\
    &Male  &1319\\
     &Other  &32\\
    \hline
        \multirow{6}{*}{ \makecell{Race / \\ Ethnicity} }
        &Caucasian&1772 \\
    &Black or African-American&105\\
    &American Indian/Alaskan Native&36\\
    &Asian or Asian-American&29\\
    &Spanish/Hispanic&11\\
    &Other&37\\
    \hline
    \multirow{6}{*}{Education }
     & High school degree&66\\
      & Some college credit, no degree&22\\
       & Associate's degree&36\\
        & Bachelor's degree&1457\\
         & Graduate's degree&353\\
          & Other&56\\
    \hline
\end{tabular}
\label{tab-demo}
\end{table*}

\end{document}